\documentclass[acmsmall,noacm,authorversion,arxiv]{acmart}
\usepackage[capitalise,noabbrev]{cleveref}
\usepackage{xspace}
\usepackage{multirow}
\usepackage{mathtools}
\usepackage{algorithm}[H]
\usepackage[noend]{algpseudocode}
\usepackage{wrapfig}

\DeclareMathOperator*{\argmax}{arg\,max}

\AtBeginDocument{%
  \providecommand\BibTeX{{%
    \normalfont B\kern-0.5em{\scshape i\kern-0.25em b}\kern-0.8em\TeX}}}

\setcopyright{rightsretained}
\acmJournal{TELO}
\acmYear{2024} \acmVolume{1} \acmNumber{1} \acmArticle{1} \acmMonth{1}\acmDOI{10.1145/3696426}

\begin{document}

\newcommand{\me}{MAP-Elites}
\newcommand{\meLong}{Multi-dimensional Archive of Phenotypic Elites}

\newcommand{\mees}{MAP-Elites-ES}
\newcommand{\meesLong}{MAP-Elites Evolutionary Stategies}

\newcommand{\pgame}{PGA-MAP-Elites}
\newcommand{\pgameLong}{Policy Gradient Assisted MAP-Elites}

\newcommand{\qdpg}{QD-PG}
\newcommand{\qdpgLong}{Quality-Diversity Policy Gradient}

\newcommand{\dcgme}{DCG-MAP-Elites}
\newcommand{\dcgmeLong}{Descriptor-Conditioned Gradients MAP-Elites}

\newcommand{\dcrlme}{DCRL-MAP-Elites}
\newcommand{\dcrlmeLong}{Descriptor-Conditioned Reinforcement Learning MAP-Elites}

\newcommand{\tdthree}{TD3}
\newcommand{\tdthreeLong}{Twin-Delayed Deep Deterministic}

\newcommand{\cmaes}{CMA-ES}
\newcommand{\cmamae}{CMA-MAE}
\newcommand{\cmamega}{CMA-MEGA}
\newcommand{\pbtme}{PBT-MAP-Elites}

\newcommand{\vic}{VIC}
\newcommand{\diayn}{DIAYN}
\newcommand{\dads}{DADS}
\newcommand{\smerl}{SMERL}

\newcommand{\obs}{s}
\newcommand{\obsSpace}{\mathcal{S}}

\newcommand{\action}{a}
\newcommand{\actionSpace}{\mathcal{A}}

\newcommand{\reward}{r}

\newcommand{\discount}{\gamma}

\newcommand{\policy}{\pi}
\newcommand{\policySpace}{\Pi}
\newcommand{\critic}{Q}

\newcommand{\replayBuffer}{\mathcal{B}}

\newcommand{\desc}{d}
\newcommand{\descSpace}{\mathcal{D}}
\newcommand{\descFunction}{D}

\newcommand{\fitness}{f}
\newcommand{\fitnessFunction}{F}

\newcommand{\archive}{\mathcal{X}}

\newcommand{\policyParams}{\psi}
\newcommand{\policyParamsSpace}{\Psi}

\newcommand{\actorParams}{\phi}
\newcommand{\actorParamsSpace}{\Phi}

\newcommand{\criticParams}{\theta}
\newcommand{\criticParamsSpace}{\Theta}

\newcommand{\gabatchsize}{b_\text{GA}}
\newcommand{\pgbatchsize}{b_\text{PG}}
\newcommand{\aibatchsize}{b_\text{AI}}
\newcommand{\aebatchsize}{b_\text{AE}}
\newcommand{\qpgbatchsize}{b_\text{QPG}}
\newcommand{\dpgbatchsize}{b_\text{DPG}}

\newcommand{\ncritic}{n}
\newcommand{\npg}{m}
\newcommand{\delay}{\Delta}
\newcommand{\defeq}{\vcentcolon=}

\title[\dcrlme{}]{Synergizing Quality-Diversity with Descriptor-Conditioned Reinforcement Learning}

\author{Maxence Faldor}
\email{m.faldor22@imperial.ac.uk}
\orcid{0000-0003-4743-9494}
\affiliation{%
  \institution{Imperial College London}
  \city{London}
  \country{United Kingdom}
}

\author{Félix Chalumeau}
\email{f.chalumeau@instadeep.com}
\orcid{0000-0001-9476-2900}
\affiliation{%
  \institution{InstaDeep}
  \city{Cape Town}
  \country{South Africa}
}

\author{Manon Flageat}
\email{manon.flageat18@imperial.ac.uk}
\orcid{0000-0002-4601-2176}
\affiliation{%
  \institution{Imperial College London}
  \city{London}
  \country{United Kingdom}
}

\author{Antoine Cully}
\email{a.cully@imperial.ac.uk}
\orcid{0000-0002-3190-7073}
\affiliation{%
  \institution{Imperial College London}
  \city{London}
  \country{United Kingdom}
}

\renewcommand{\shortauthors}{Faldor, et al.}

\begin{abstract}
A hallmark of intelligence is the ability to exhibit a wide range of effective behaviors. Inspired by this principle, Quality-Diversity algorithms, such as \me{}, are evolutionary methods designed to generate a set of diverse and high-fitness solutions.
However, as a genetic algorithm, \me{} relies on random mutations, which can become inefficient in high-dimensional search spaces, thus limiting its scalability to more complex domains, such as learning to control agents directly from high-dimensional inputs.
To address this limitation, advanced methods like \pgame{} and \dcgme{} have been developed, which combine actor-critic techniques from Reinforcement Learning with \me{}, significantly enhancing the performance and efficiency of Quality-Diversity algorithms in complex, high-dimensional tasks.
While these methods have successfully leveraged the trained critic to guide more effective mutations, the potential of the trained actor remains underutilized in improving both the quality and diversity of the evolved population.
In this work, we introduce \dcrlme{}, an extension of \dcgme{} that utilizes the descriptor-conditioned actor as a generative model to produce diverse solutions, which are then injected into the offspring batch at each generation.
Additionally, we present an empirical analysis of the fitness and descriptor reproducibility of the solutions discovered by each algorithm.
Finally, we present a second empirical analysis shedding light on the synergies between the different variations operators and explaining the performance improvement from \pgame{} to \dcrlme{}.
\end{abstract}





\begin{teaserfigure}
\vspace{-0.2cm}
\centering
\includegraphics[width=0.7\textwidth]{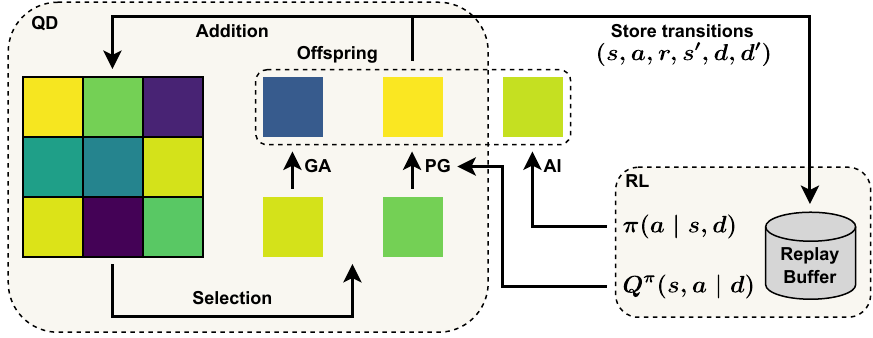}
\vspace{-0.2cm}
\caption{
\dcrlme{} employs a standard Quality-Diversity loop comprising selection, variation, evaluation and addition.
Concurrently, transitions generated during the evaluation step are stored in a replay buffer and used to train a descriptor-conditioned actor-critic model from reinforcement learning.
Two complementary variation operators are used: a Genetic Algorithm (\textbf{GA}) variation operator for diversity and a Policy Gradient (\textbf{PG}) variation operator for quality.
During the actor-critic training, the diverse and high-performing policies from the archive are distilled into the generally capable actor.
In turn, this descriptor-conditioned actor is utilized as a generative model to produce diverse solutions, which are then injected (\textbf{AI}) into the offspring batch at each generation.
}
\Description{\dcrlme{} diagram.}
\label{fig:teaser}
\end{teaserfigure}

\maketitle

\newcommand{\MyLabel}[1]{\label{\MyTag#1}}
\newcommand{\MyRef}[1]{\cref{\MyTag#1}}

\begingroup
\let\clearpage\relax
\newcommand{\MyTag}{main}
\section{Introduction}
\label{sec:introduction}
A remarkable feature of biological evolution is its capacity to produce a diverse array of species, each uniquely adapted to its ecological niche.
Inspired by this phenomenon, Quality-Diversity (QD) is a family of evolutionary methods that aims to emulate nature's inventiveness by generating diverse populations of high-fitness individuals~\citep{pugh_QualityDiversityNew_2016,cully_QualityDiversityOptimization_2017,chatzilygeroudis_QualityDiversityOptimizationNovel_2021}.
In contrast with traditional optimization methods that focus only on finding the global optimum and return a single solution, the goal of QD algorithms is to illuminate a search space of interest, known as the \emph{descriptor space}, by discovering a variety of high-performing solutions across different niches~\citep{mouret_IlluminatingSearchSpaces_2015}.
QD algorithms have demonstrated promising results in improving exploration~\citep{ecoffet_FirstReturnThen_2021}, enhancing robustness and adaptability~\citep{cully_RobotsThatCan_2015}, reducing the reality gap~\citep{chatzilygeroudis_ResetfreeTrialandErrorLearning_2018}, and fostering creativity~\citep{faldor2024leniabreeder}.
The generation of a diverse collection of effective solutions provides multiple alternatives for solving a single problem, proving particularly valuable in robotics for improving robustness and adaptability~\citep{cully_RobotsThatCan_2015,airl2024qdac}.
Moreover, while gradient-based optimization methods often struggle with local optima, maintaining a repertoire of diverse solutions can reveal stepping stones towards globally superior solutions~\citep{mouret_IlluminatingSearchSpaces_2015,nilsson_PolicyGradientAssisted_2021}.

Among QD algorithms, \meLong{} (\me{}) stands out as a conceptually simple yet highly effective method~\citep{mouret_IlluminatingSearchSpaces_2015}.
However, MAP-Elites primarily relies on random mutations for exploration. This reliance can become inefficient in high-dimensional search spaces, potentially limiting its scalability to more complex domains, such as learning to control agents directly from high-dimensional inputs~\citep{nilsson_PolicyGradientAssisted_2021}.

Reinforcement Learning (RL) has achieved remarkable milestones across diverse domains, from mastering discrete games~\citep{mnih_PlayingAtariDeep_2013,silver_MasteringGameGo_2016} to solving continuous control challenges in locomotion~\citep{haarnoja_SoftActorCriticAlgorithms_2019,heess_EmergenceLocomotionBehaviours_2017} and manipulation~\citep{openai_SolvingRubikCube_2019}.
These accomplishments have demonstrated the exceptional capability of RL algorithms to solve complex, specialized tasks.
In particular, policy gradient methods have shown state-of-the-art results in learning large neural network policies with thousands of parameters in high-dimensional and continuous domains~\citep{lillicrap_ContinuousControlDeep_2019,silver_DeterministicPolicyGradient_2014,haarnoja_SoftActorCriticOffPolicy_2018}.
Recognizing the complementary strengths of QD and RL, numerous methods have been proposed to combine these approaches, aiming to enhance the performance of QD algorithms in complex, high-dimensional tasks while maintaining their ability to generate diverse policies~\citep{nilsson_PolicyGradientAssisted_2021,pierrot_DiversityPolicyGradient_2022,tjanaka_ApproximatingGradientsDifferentiable_2022}.

Among these hybrid approaches, \pgameLong{} (\pgame{}) has emerged as a particularly promising method, achieving strong performance on challenging continuous control locomotion tasks~\cite{nilsson_PolicyGradientAssisted_2021}.
\pgame{} integrates \me{} with RL, leveraging the \tdthreeLong{} (\tdthree{}) policy gradient algorithm to enhance sample efficiency~\citep{fujimoto_AddressingFunctionApproximation_2018}.
The algorithm maintains a standard QD loop of selection, variation, evaluation, and addition, while concurrently storing generated transitions in a replay buffer.
These transitions are used to train an actor-critic model using \tdthree{}.
\pgame{} employs two complementary variation operators: a Genetic Algorithm (GA) variation for exploration, coupled with a Policy Gradient (PG) variation for fitness improvement, that leverages gradients derived from the trained critic using RL.
However, \pgame{} faces a limitation stemming from the inherent tension between the convergent nature of RL and the divergent search of QD. In tasks where the global optimum identified by the RL algorithm fails to generate diverse, high-performing solutions, the PG variation becomes ineffective.

\dcgmeLong{} (\dcgme{}) overcomes the limitations of \pgame{} by leverage a descriptor-conditioned RL algorithm, achieving state-of-the-art performance on challenging continuous control locomotion tasks~\citep{faldor_dcgme_2023}.
\dcgme{} follows the same overall structure as \pgame{}, but replaces the standard actor-critic model with a descriptor-conditioned variant.
The descriptor-conditioned critic guides policy gradient updates by providing value estimates tailored to specific target descriptors. This mechanism allows the PG operator to mutate solutions, producing offspring with higher fitness while maintaining their original descriptors.
As a by-product of the actor-critic training, the diverse, high-performing policies from the archive are distilled into the descriptor-conditioned actor.
This resulting actor, conditionable on any descriptor from the archive, emerges as a versatile agent demonstrating a wide range of behaviors.

A key challenge in \dcgme{} is reconciling the mismatch between the unconditioned policies in the archive and the descriptor-conditioned nature of the actor-critic algorithm.
Indeed, the actor-critic algorithm is trained on the transitions collected from evaluating these unconditioned policies, which only provide observed descriptors but lack the crucial target descriptors required for training a descriptor-conditioned RL algorithm.
To overcome this limitation and stabilize the actor-critic training, \dcgme{} evaluates the actor itself, storing the generated transitions in the replay buffer.
This strategy serves two crucial purposes. First, it generates active samples, known to enhance training stability~\citep{ostrovski_difficulty_2021}.
Second, it produces transitions containing both target and observed descriptors, thus providing the necessary data for effectively training the descriptor-conditioned actor-critic model.
However, this actor evaluation is not cheap as it contributes to reducing the sample efficiency of the algorithm.
Moreover, while \dcgme{} has successfully leveraged the trained critic to guide more effective mutations, the potential of the trained actor remains underutilized in improving both the quality and diversity of the evolved population.

In this work, we introduce \dcrlme{}, an extension of \dcgme{} that fully leverages both components of the RL algorithm. While \dcgme{} primarily utilized the critic, \dcrlme{} also harnesses the descriptor-conditioned actor as a generative model to produce diverse policies, which are then injected into the offspring batch at each generation.
This approach, named Actor Injection (AI), significantly enhances both the quality and diversity of the archive.
By leveraging the actor's learned representation of the descriptor space, \dcrlme{} can generate policies tailored to specific niches, potentially discovering high-performing solutions in unexplored or underexplored regions.
This novel mechanism inserts the descriptor-conditioned actor learned with reinforcement learning within the population, despite differences in policy architectures, introducing novel genetic material, promoting diversity and reducing the risk of premature convergence.
Moreover, this method elegantly addresses the issue of unconditioned transitions faced by \dcgme{}.
Since the actor is conditioned on descriptors, the generated policies inherently contain both target and observed descriptors, eliminating the need for separate actor evaluation.
This not only solves the mismatch problem but also significantly improves sample efficiency by reducing the number of environment interactions required.
By leveraging the actor as a generative model, \dcrlme{} effectively combines the exploration capabilities of QD algorithms with the efficient learning of RL, resulting in a more powerful and sample-efficient algorithm for discovering diverse, high-quality solutions in complex control tasks.

We compare our algorithm to five QD algorithms across seven challenging continuous control locomotion tasks. Our method, \dcrlme{}, consistently achieves equal or higher QD scores and coverage compared to all baselines across all tasks, demonstrating its superior performance in generating diverse, high-quality solutions.
To provide a deeper understanding of \dcrlme{} and \dcgme{}'s capabilities, we conduct two empirical analyses. First, we evaluate the fitness and descriptor reproducibility of the solutions discovered by each algorithm in the face of environment stochasticity. This analysis assesses their variance across multiple evaluations, offering insights into the reproducibility of the learned policies.
Second, we present an empirical study that illuminates the synergies between the different variation operators employed in our method. This analysis traces the improvements from \pgame{} to our proposed method \dcrlme{}, providing a clear picture of how each component contributes to the overall performance gain.

\section{Background}
\label{sec:background}
\subsection{Problem Statement}
We consider the reinforcement learning framework~\cite{sutton_ReinforcementLearningIntroduction_2018} where an agent sequentially interacts with an environment at discrete time steps $t$, modeled as a \emph{Markov Decision Process} (MDP).
At each time step $t$, the agent observes a state $\obs_t \in \obsSpace$, takes an action $\action_t \in \actionSpace$ and receives a scalar reward $\reward_t = r(\obs_t, \action_t) \in \mathbb{R}$. The action cause the environment to transition to a next state $\obs_{t+1} \in \obsSpace$, sampled from the dynamics $p(\obs_{t+1} \mid \obs_t, \action_t)$.
The agent uses its \emph{policy} to select actions and interacts with the environment to give a trajectory of states, actions and rewards.
In this work, we only consider deterministic policies. A policy with parameters $\actorParams$ is denoted $\policy_\actorParams \in \policySpace$ and is a mapping from state to action $\policy_\actorParams \colon \obsSpace \to \actionSpace$.

QD algorithms aim to discover a diverse set of high-performing solutions.
In the context of RL, the \emph{fitness function} $F \colon \policySpace \to \mathbb{R}$ quantifies the performance of policies and is typically defined as the expected sum of rewards over an episode of length $T$, $\mathbb{E}_{\policy} \left[ \sum_{t=0}^{T-1} \reward_t \right]$.
The \emph{descriptor function} $D \colon \policySpace \to \descSpace \subset \mathbb{R}^d$ maps policies to a descriptor space $\descSpace$ and is generally defined by the user to characterize solutions in a meaningful way for the type of diversity desired.
In this setting, the objective of QD algorithms is to find the highest fitness solutions in each point of the \emph{descriptor space} $\descSpace$. With these notations, our objective is to evolve a population of solutions that are both high-performing with respect to $F$ and diverse with respect to $D$.

\subsection{\me{}}
\me{}~\citep{mouret_IlluminatingSearchSpaces_2015} is a simple yet effective QD algorithm, that discretizes the descriptor space $\descSpace$ into a multi-dimensional grid of cells called archive $\archive$ and searches for the best solution in each cell, see \cref{alg:me}.
The goal of the algorithm is to return an archive that is filled as much as possible with high-fitness solutions.
\me{} starts by initializing the archive with randomly generated solutions. The algorithm then repeats the following steps until a budget of $I$ solutions have been evaluated: (1) a batch of solutions from the archive are uniformly selected and modified through mutations and/or crossovers to produce offspring, (2) the fitnesses and descriptors of the offspring are evaluated, and each offspring is placed in its corresponding cell if and only if the cell is empty or if the offspring has a better fitness than the current solution in that cell, in which case the current solution is replaced.
As most evolutionary methods, \me{} relies on undirected updates that are agnostic to the fitness objective. With a Genetic Algorithm (GA) variation operator, \me{} performs a divergent search that may cause slow convergence in high-dimensional problems due to a lack of directed search power, and thus, is performing best on low-dimensional search space~\citep{nilsson_PolicyGradientAssisted_2021}.

\subsection{Deep Reinforcement Learning}
\label{sec:background-drl}
Deep Reinforcement Learning~\citep{mnih_HumanlevelControlDeep_2015} combines the reinforcement learning framework with the function approximation capabilities of deep neural networks to represent policies and value functions in high-dimensional state and action spaces. In opposition to black-box optimization methods like evolutionary algorithms, RL leverages the structure of the MDP in the form of the Bellman equation to achieve better sample efficiency. The objective is to find an optimal policy $\policy_\actorParams$, which maximizes the expected return or fitness $F(\policy_\actorParams)$. In reinforcement learning, many approaches try to estimate the action-value function $\critic^\policy(\obs, \action) = \mathbb{E}_\policy \left[ \sum_{i=0}^{T-t-1} \discount^i \reward_{t+i} \mid \obs_t = \obs, \action_t = \action \right]$ defined as the expected discounted return starting from state $\obs$, taking action $\action$ and thereafter following policy $\policy$.

\tdthree{} algorithm~\citep{fujimoto_AddressingFunctionApproximation_2018} is an off-policy actor-critic reinforcement learning method that achieves state-of-the-art results in environments with large and continuous action space. \tdthree{} indirectly learns a policy $\policy_\actorParams$ via maximization of the action-value function $\critic_\criticParams(\obs, \action)$. The approach is closely connected to $Q$-learning~\citep{fujimoto_AddressingFunctionApproximation_2018} and tries to approximate the optimal action-value function $\critic^*(\obs, \action)$ in order to find the optimal action $\policy^*(\obs) = \argmax_\action \critic^*(\obs, \action)$. However, computing the maximum over action in $\max_\action \critic_\criticParams(\obs, \action)$ is intractable in continuous action space, hence it is approximated using $\max_\action \critic_\criticParams(\obs, \action) = \critic_\criticParams(\obs, \policy_\actorParams(\obs))$. In \tdthree{}, the policy $\policy_\actorParams$ takes actions in the environment and the transitions are stored in a replay buffer. The collected experience is then used to train a pair of critics $\critic_{\criticParams_1}$, $\critic_{\criticParams_2}$ using temporal difference. Target networks $\critic_{{\criticParams_1}^\prime}$, $\critic_{{\criticParams_2}^\prime}$ are updated to slowly track the main networks. Both critics use a single regression target $y$, calculated using whichever of the two target critics gives a smaller estimated value and using target policy smoothing by sampling a noise $\epsilon \sim \text{clip}(\mathcal{N}(0, \sigma), -c, c)$:
\begin{equation}
\label{eq:td3-target}
y = \reward(\obs_t, \action_t) + \discount \min_{i=1,2} \critic_{{\criticParams_i}^\prime}(\obs_{t+1}, \policy_{\actorParams^\prime}(\obs_{t+1}) + \epsilon)
\end{equation}
Both critics are learned by regression to this target and the policy is learned with a delay, only updated every $\delay$ iterations simply by maximizing $\critic_{\criticParams_1}$ with $\max_\actorParams \mathbb{E} \left[ \critic_{\criticParams_1} (\obs, \policy_\actorParams(\obs)) \right]$. The actor is updated using the deterministic policy gradient:
\begin{equation}
\label{eq:pg}
\nabla_\actorParams J(\actorParams) = \mathbb{E} \left[ \nabla_\actorParams \policy_{\actorParams}(\obs) \nabla_\action \critic_{\criticParams_1}(\obs, \action)|_{\action=\policy_{\actorParams}(\obs)} \right]
\end{equation}

\subsection{\pgame{}}
\label{sec:background-pga-me}
\pgame{}~\citep{nilsson_PolicyGradientAssisted_2021} is an extension of \me{} that is designed to evolve deep neural networks by combining the directed search power and sample efficiency of RL methods with the exploration capabilities of genetic algorithms, see \cref{alg:pga-me}. The algorithm follows the usual \me{} loop of selection, variation, evaluation and addition for a budget of $I$ iterations, but uses two parallel variation operators: half of the offspring are generated using a standard Genetic Algorithm (GA) variation operator and half of the offspring are generated using a Policy Gradient (PG) variation operator. During each iteration of the loop, \pgame{} stores the transitions from offspring evaluation in a replay buffer $\replayBuffer$ and uses it to train a pair of critics based on the \tdthree{} algorithm, described in \cref{alg:pga-me-actor-critic}. The trained critic is then used in the PG variation operator to update the selected solutions from the archive for $\npg$ gradient steps to select actions that maximize the approximated action-value function, as described in \cref{alg:pga-me-qpg}. At each iteration, the critics are trained for $\ncritic$ steps of gradients descents towards the target described in \cref{eq:td3-target}, averaged over $N$ transitions of experience sampled uniformly from the replay buffer $\replayBuffer$. The actor learns with a delay $\delay$ via maximization of the critic according to \cref{eq:pg}.

\section{Methods}
\label{sec:methods}
\begin{algorithm}
\caption{\dcrlme{}}
\MyLabel{alg:dcrl-me}
\begin{algorithmic}
\Require GA batch size $\gabatchsize$, PG batch size $\pgbatchsize$, Actor Injection batch size $\aibatchsize$, total batch size $b = \gabatchsize + \pgbatchsize + \aibatchsize$
\State Initialize archive $\archive$ with $b$ random solutions and replay buffer $\replayBuffer$
\State Initialize critic networks $\critic_{\criticParams_1}$, $\critic_{\criticParams_2}$ and actor network $\policy_\actorParams$
\State $i \gets 0$
\While{$i < I$}
    \State $\textsc{train\_actor\_critic}(\policy_\actorParams, \critic_{\criticParams_1}, \critic_{\criticParams_2}, \replayBuffer)$
    \State $\policy_{\policyParams_1}, \dots, \policy_{\policyParams_b} \gets \textsc{selection}(\archive)$
    \State $\policy_{\widehat{\policyParams}_1}, \dots, \policy_{\widehat{\policyParams}_{\gabatchsize}} \gets \textsc{variation\_ga}(\policy_{\policyParams_1}, \dots, \policy_{\policyParams_{\gabatchsize}})$
    \State $\policy_{\widehat{\policyParams}_{\gabatchsize+1}}, \dots, \policy_{\widehat{\policyParams}_{\gabatchsize+\pgbatchsize}} \gets \textsc{variation\_pg}(\policy_{\policyParams_{\gabatchsize+1}}, \dots, \policy_{\policyParams_{\gabatchsize+\pgbatchsize}}, \critic_{\criticParams_1}, \replayBuffer)$
    \State $\policy_{\widehat{\policyParams}_{\gabatchsize+\pgbatchsize+1}}, \dots, \policy_{\widehat{\policyParams}_b} \gets \textsc{actor\_injection}(\policy_\actorParams)$
    \State $\textsc{addition}(\policy_{\widehat{\policyParams}_1}, \dots, \policy_{\widehat{\policyParams}_b}, \archive, \replayBuffer)$
    \State $i\gets i + b$  
\EndWhile

\Function{\textsc{addition}}{$\policy_{\widehat{\policyParams}} \dots, \archive, \replayBuffer$}
    \For{$\policy_{\widehat{\policyParams}} \dots$}
        \State $(f, \text{transitions}) \gets F(\policy_{\widehat{\policyParams}})$, $\desc \gets D(\policy_{\widehat{\policyParams}})$
        \State $\textsc{insert}(\replayBuffer, \text{transitions})$
        \If{$\archive(\desc) = \emptyset$ or $F(\archive(\desc)) < f$}
            \State $\archive(\desc) \gets \policy_{\widehat{\policyParams}}$
        \EndIf
    \EndFor
\EndFunction
\end{algorithmic}
\end{algorithm}

Our method, \dcrlmeLong{} (\dcrlme{}), extends \dcgme{} by utilizing the descriptor-conditioned actor as a generative model to produce diverse policies, which are then injected into the offspring batch at each generation. Like \dcgme{}, our method leverages a descriptor-conditioned actor-critic model that enhances the PG variation operator, while simultaneously distilling the archive into a single policy. The pseudocode is provided in \MyRef{alg:dcrl-me}. \dcrlme{} employs a standard QD loop comprising selection, variation, evaluation and addition for a budget of $I$ iterations (\cref{fig:teaser}).
Concurrently, transitions generated during the evaluation step are stored in a replay buffer and used to train a descriptor-conditioned actor-critic model using reinforcement learning.

Contrary to \pgame{}, the actor-critic pair is descriptor-conditioned in \dcgme{} and \dcrlme{}. In addition to the state $\obs$ and action $\action$, the critic $\critic_\criticParams\left(\obs, \action \mid \desc\right)$ also depends on the descriptor $\desc$ and estimates the expected discounted return starting from state $\obs$, taking action $\action$ and thereafter following policy $\policy$ \emph{and} achieving descriptor $\desc$.
In this work, to achieve descriptor $\desc$ means that the trajectory generated by the policy $\policy$ has descriptor $\desc$. In addition to the state $\obs$, the actor $\policy_\actorParams(\obs \mid \desc)$ also depends on a target descriptor $\desc$ and maximizes the expected discounted return conditioned on achieving the target descriptor $\desc$. Thus, the goal of the descriptor-conditioned actor is to achieve the desired descriptor $\desc$ \emph{while} maximizing fitness.

The descriptor-conditioned critic provides action-value estimates tailored to specific target descriptors (\cref{sec:methods-dc-critic}).
As a by-product of the actor-critic training, the diverse, high-performing policies from the archive are distilled into the descriptor-conditioned actor. This resulting actor, conditionable on any descriptor from the archive, emerges as a generally capable agent demonstrating a wide range of behaviors. (\cref{sec:methods-dc-actor}).
The descriptor-conditioned actor-critic is trained following \tdthree{} algorithm (\cref{sec:methods-actor-critic-training}).
The PG variation operator utilizes the descriptor-conditioned critic to mutate policies stored in the archive, producing offspring with higher fitness while maintaining their original descriptors (\cref{sec:methods-dc-pg}).
Finally, \dcrlme{} extends \dcgme{} by using the descriptor-conditioned actor as a generative model to produce diverse policies, which are then injected into the offspring batch at each generation (\cref{sec:methods-actor-injection}). By leveraging the actor’s learned representation of the descriptor space, \dcrlme{} can generate policies tailored to specific niches, potentially discovering high-performing solutions in unexplored or underexplored regions.

\subsection{Descriptor-Conditioned Critic}
\label{sec:methods-dc-critic}
Instead of estimating the action-value function with $\critic_\criticParams(\obs, \action)$, we want to estimate the descriptor-conditioned action-value function with $\critic_\criticParams(\obs, \action \mid \desc)$.
When a policy $\policy$ interacts with the environment, it generates a trajectory, which is a sequence of transitions $\left(s, a, r, s^\prime\right)$ with descriptor $\desc$. We extend the definition of a transition $(\obs, \action, \reward, \obs^\prime)$ to include the observed descriptor $\desc$ of the trajectory $(\obs, \action, \reward, \obs^\prime, \desc)$. However, the descriptor is only available at the end of the episode, therefore the transitions can only be augmented with the descriptor after the episode is completed.
In all the tasks we consider, the reward function is positive $\reward \colon \mathcal{S} \times \mathcal{A} \to \mathbb{R}^+$ and hence, the fitness function $F$ and action-value function are positive as well. Thus, for any target descriptor $\desc^\prime \in \mathcal{D}$, we define the descriptor-conditioned critic as equal to the normal action-value function when the policy achieves the target descriptor $\desc^\prime$ and as equal to zero when the policy does not achieve the target descriptor $\desc^\prime$. Given a transition $(\obs, \action, \reward, \obs^\prime, \desc)$, and a target descriptor $\desc^\prime$ sampled in $\mathcal{D}$,
\begin{equation}
\label{eq:dc-critic-1}
\critic_\criticParams(\obs, \action \mid \desc^\prime) \defeq 
    \begin{cases}
        \critic_\criticParams(\obs, \action), & \text{if } \desc = \desc^\prime\\
        0, & \text{if } \desc \neq \desc^\prime
    \end{cases}
\end{equation}
However, with this piecewise definition, the descriptor-conditioned action-value function is not continuous and violates the universal approximation theorem continuity hypothesis~\citep{hornik_MultilayerFeedforwardNetworks_1989}. To address this issue, we replace the piecewise definition with a smooth similarity function $S \colon \mathcal{D}^2 \to ]0, 1]$ defined as $S(\desc, \desc^\prime) = \exp(-||\desc-\desc^\prime||_\mathcal{D}/L)$. This similarity function gives a measure of how close two descriptors are to each other, and is controlled by a length scale parameter $L$. Notice that, as the length scale $L$ tends towards $0$, we approach the initial piecewise definition. Therefore, we relax the definition of the descriptor-conditioned critic given in \cref{eq:dc-critic-1} into:
\begin{align}
\label{eq:dc-critic-2}
\critic_\criticParams \left( \obs, \action  \mid \desc^\prime \right) = S(\desc, \desc^\prime) \, \critic_\criticParams(\obs, \action)\nonumber &= S(\desc, \desc^\prime) \, \mathbb{E}_\policy \left[ \sum_{i=0}^{T-t-1} \discount^i \reward_{t+i} \middle| \obs, \action \right]\nonumber\\
&= \mathbb{E}_\policy \left[ \sum_{i=0}^{T-t-1} \discount^i S(\desc, \desc^\prime) \reward_{t+i} \middle| \obs, \action \right]
\end{align}
With \cref{eq:dc-critic-2}, we demonstrate that learning the descriptor-conditioned critic is equivalent to scaling the reward by the similarity $S(\desc, \desc^\prime)$ between the descriptor of the trajectory $\desc$ and the target descriptor $\desc^\prime$. Therefore, the critic target in \cref{eq:td3-target} is modified to include the similarity scaling and the descriptor-conditioned actor:
\begin{equation}
\label{eq:dc-td3-target}
y = S(\desc, \desc^\prime) \, \reward(\obs_t, \action_t) + \discount \min_{i=1,2} \critic_{{\criticParams_i}^\prime} (\obs_{t+1}, \policy_{\actorParams^\prime}(\obs_{t+1} \mid \desc^\prime) + \epsilon \mid \desc^\prime)
\end{equation}
If the target descriptor $\desc^\prime$ is approximately equal to the observed descriptor $\desc$ of the trajectory $\desc \approx \desc^\prime$, then we have $S(\desc, \desc^\prime) \approx 1$ so the reward is unchanged. However, if the descriptor $\desc^\prime$ is different from the observed descriptor $\desc$, then the
reward is scaled down to $S(\desc, \desc^\prime) \, \reward(\obs_t, \action_t) \approx 0$. The scaling ensures that the magnitude of the reward depends not only on the quality of the action $\action$ with regards to the fitness function $F$, but also on achieving the target descriptor $\desc^\prime$. Given one transition $(\obs, \action, \reward, \obs^\prime, \desc)$, we can generate infinitely many critic updates by sampling a target descriptor $\desc^\prime \in \mathcal{D}$. This is leveraged in the new actor-critic training introduced with \dcgme{}, which is detailed in \MyRef{alg:dcrl-me-actor-critic} and \cref{sec:methods-actor-critic-training}.

\subsection{Descriptor-Conditioned Actor and Archive~Distillation}
\label{sec:methods-dc-actor}
As explained in \cref{sec:background-drl}, the training of the critic requires to train an actor $\policy_\actorParams$ to approximate the optimal action $\action^*$. However, in this work, the action-value function estimated by the critic is conditioned on a descriptor $\desc$. Hence, we don't want $\policy_\actorParams$ to estimate the best action globally, but rather the best action given that it achieves the target descriptor $\desc$. Therefore, the actor is extended to a descriptor-conditioned policy $\policy_\actorParams(\obs \mid \desc)$, that maximizes the descriptor-conditioned critic's value with $\max_\actorParams \mathbb{E} \left[ \critic_\criticParams (\obs, \policy_\actorParams (\obs \mid \desc) \mid \desc) \right]$. The actor is updated using the deterministic policy gradient, see \MyRef{alg:dcrl-me-actor-critic}:
\begin{equation}
\label{eq:dc-pg-actor}
\nabla_\actorParams J(\actorParams) = \frac{1}{N}\sum \nabla_\actorParams \policy_{\actorParams}(\obs \mid \desc^\prime) \nabla_\action \critic_{\criticParams_1}(\obs, \action \mid \desc^\prime)|_{\action=\policy_{\actorParams}(\obs \mid \desc^\prime)}
\end{equation}
The policy $\policy_\actorParams(\obs \mid \desc)$ learns to suggest actions $\action$ that optimize the return \emph{while} generating a trajectory achieving descriptor $\desc$. Consequently, the descriptor-conditioned actor can exhibit a wide range of descriptors, effectively distilling some of the capabilities of the archive into a single versatile policy.

\subsection{Actor-Critic Training}
\label{sec:methods-actor-critic-training}
\begin{algorithm}[H]
\caption{Descriptor-Conditioned Actor-Critic Training}
\MyLabel{alg:dcrl-me-actor-critic}
\begin{algorithmic}
\Function{\textsc{train\_actor\_critic}}{$\policy_\actorParams, \critic_{\criticParams_1}, \critic_{\criticParams_2}, \replayBuffer$}
    \For{$t = 1  \rightarrow \ncritic$}
        \State Sample $N$ transitions $\left(\obs, \action, \reward, \obs^\prime, \desc, \desc^\prime \right)$ from $\replayBuffer$
        \State Sample smoothing noise $\epsilon$
        \State $y \gets S(\desc, \desc^\prime) \, \reward + \discount \min\limits_{i=1,2}  \critic_{\criticParams_{i}^\prime}(\obs^\prime, \policy_{\actorParams^\prime}(\obs^\prime \mid \desc^\prime) + \epsilon \mid \desc^\prime)$
        \State Update both critics by regression to $y$
        \If{$t$ mod $\delay$}
            \State Update actor using the deterministic policy gradient:
            \State 	$\frac{1}{N}\sum \nabla_\actorParams \policy_{\actorParams}(\obs \mid \desc^\prime) \nabla_\action \critic_{\criticParams_1}(\obs, \action \mid \desc^\prime)|_{\action=\policy_{\actorParams}(\obs \mid \desc^\prime)}$
            \State Soft-update target networks $\critic_{\criticParams{i}^\prime}$ and $\policy_{\actorParams^\prime}$
        \EndIf
    \EndFor
\EndFunction
\end{algorithmic}
\end{algorithm}

In \cref{sec:methods-dc-critic}, we show that the descriptor-conditioned critic target $y$ in \cref{eq:dc-td3-target} requires a transition $(\obs, \action, \reward, \obs^\prime, \desc)$ and a target descriptor $\desc^\prime$.
Most related methods that are conditioned on skills or goals rely on a sampling strategy. For example, HER~\citep{andrychowicz_HindsightExperienceReplay_2017} is a goal-conditioned reinforcement learning algorithm that relies on a handcrafted goal sampling strategy and \diayn{}, \dads{}, \smerl{} sample skills from a uniform prior distribution. However, in this work, we don't need to rely on an explicit descriptor sampling strategy.

For each PG variation operator offspring, the transitions coming from the evaluation step, are populated with $\desc^\prime$ equal to the descriptor of the parent solution $\desc_\policyParams$. The PG variation operator mutates the parent to improve fitness while achieving descriptor $\desc_\policyParams$. Thus, although the offspring is not descriptor-conditioned, its implicit target descriptor is $\desc_\policyParams$. Consequently, we set the target descriptor $\desc^\prime$ to the descriptor of the parent $\desc_\policyParams$.

Similarly, for each GA variation operator offspring, the transitions coming from the evaluation step, are populated with $\desc^\prime$ equal to the observed descriptor of the trajectory $\desc$. The GA variation operator mutates the parent by adding random noise to the genotype. However, a small random change in the parameters of the parent solution can induce big changes in the behavior of the offspring, making them behaviorally different. Consequently, we set the target descriptor $\desc^\prime$ to the observed descriptor of the trajectory $\desc$.

At the end of the evaluation step, we augment the transitions with the observed descriptor of the trajectory $\desc$, and with the target descriptor $\desc^\prime$, using the implicit descriptor sampling strategy explained above, giving $\left(\obs, \action, \reward, \obs^\prime, \desc, \desc^\prime \right)$.
This implicit descriptor sampling strategy has two benefits. First, half of the transitions have $\desc = \desc^\prime$, providing the actor-critic training with samples where the target descriptor is achieved, therefore alleviating sparse reward problems.
Second, at the beginning of the training process, half of the transitions will have $\desc \neq \desc^\prime$ because the solutions in the archive have not learned to accurately achieve their descriptors yet. However, as training goes on, the number of samples where the descriptor is not achieved will decrease, providing some kind of automatic curriculum. Finally, the actor-critic training is adapted from \tdthree{} and is given in \MyRef{alg:dcrl-me-actor-critic}.

\subsection{Descriptor-Conditioned PG Variation}
\label{sec:methods-dc-pg}
\begin{algorithm}[H]
\caption{Descriptor-Conditioned PG Variation}
\MyLabel{alg:dcrl-me-qpg}
\begin{algorithmic}
\Function{\textsc{variation\_pg}}{$\policy_{\policyParams} \dots,  \critic_{\criticParams_1}, \replayBuffer$}
    \For{$\policy_{\policyParams} \dots$}
        \State $\desc_\policyParams \gets \descFunction(\policy_{\policyParams})$
        \For{$i = 1  \rightarrow \npg$}
            \State Sample $N$ transitions $\left(\obs, \action, \reward, \obs^\prime, \desc, \desc^\prime \right)$ from $\replayBuffer$
            \State Update actor using the deterministic policy gradient:
            \State 	$\frac{1}{N}\sum \nabla_\policyParams \policy_{\policyParams}(\obs) \nabla_\action \critic_{\criticParams_1}(\obs, \action \mid \desc_\policyParams)|_{\action=\policy_{\policyParams}(\obs)}$
        \EndFor
    \EndFor
    \State \Return $\policy_{\widehat{\actorParams}} \dots$
\EndFunction
\end{algorithmic}
\end{algorithm}

Once the critic $\critic_\criticParams(\obs, \action \mid \desc)$ is trained, it can be used to improve the fitness of any solutions in the archive, as described in \MyRef{alg:dcrl-me-qpg}.
First, a parent solution $\policy_\policyParams$ is selected from the archive and we denote its descriptor by $\desc_\policyParams \defeq D(\policy_\policyParams)$. Notice that this policy $\policy_\policyParams(\obs)$ is not descriptor-conditioned, contrary to the actor $\policy_\actorParams(\obs \mid \desc)$.
Second, we apply the PG variation operator from \cref{eq:dc-pg}, for $\npg$ gradient steps, using the descriptor $\desc_\policyParams$ to condition the critic:
\begin{equation}
\label{eq:dc-pg}
\nabla_\policyParams J(\policyParams) = \frac{1}{N}\sum \nabla_\policyParams \policy_{\policyParams}(\obs) \nabla_\action \critic_{\criticParams_1}(\obs, \action \mid \desc_\policyParams)|_{\action=\policy_{\policyParams}(\obs)}
\end{equation}
The goal is to improve the quality of the solution $\policy_\policyParams$, while keeping the same diversity $\desc_\policyParams$. To that end, the critic is used to evaluate actions and guides $\policy_\policyParams$ to (1) improve fitness, while (2) achieving descriptor $\desc_\policyParams$.

\subsection{Descriptor-Conditioned Actor Injection}
\label{sec:methods-actor-injection}
\newcommand{\weight}{\mathbf{W}}
\newcommand{\bias}{\mathbf{b}}
\begin{algorithm}[H]
\caption{Descriptor-Conditioned Actor Injection}
\MyLabel{alg:dcrl-me-ai}
\begin{algorithmic}
\Function{\textsc{actor\_injection}}{$\policy_\actorParams$}
    \State $\desc_1, \dots, \desc_{\aibatchsize} \sim \mathcal{U}(\descSpace)$
    \State $\policyParams_1, \dots, \policyParams_{\aibatchsize} \gets G_\actorParams(\desc_1), \dots, G_\actorParams(\desc_{\aibatchsize})$
    \State \Return $\policy_{\policyParams_1}, \dots, \policy_{\policyParams_{\aibatchsize}}$
\EndFunction
\end{algorithmic}
\end{algorithm}

In QD algorithms, such as \pgame{}, the addition of the trained actor into the archive of solutions, a process known as Actor Injection (AI), has been shown to significantly improve performance~\cite{flageat_EmpiricalAnalysisPGAMAPElites_2023}.
However, our descriptor-conditioned approach presents a unique challenge: the architecture of the descriptor-conditioned actor does not match that of the policies stored in the archive.
This mismatch arises because the policies in the archive are designed to take only the state as input, whereas our descriptor-conditioned actor accepts both state and descriptor as inputs.

The core of our approach lies in the creation of ``specialized'' versions of the generally capable actor.
We begin by uniformly sampling a set of descriptors from the descriptor space.
For each sampled descriptor, we generate a specialized version of the actor that is tailored to work with that specific descriptor.
Crucially, we then transform these specialized versions to match the architecture of the policies in the archive.
This transformation is achieved by extracting the weights corresponding to the state input and adjusting the biases to incorporate the effect of the fixed descriptor.
This technique allows us to create and inject multiple specialized versions of our versatile actor into different niches of the archive, potentially improving performance across various areas of the solution space without resorting to computationally expensive policy gradient variations.

\begin{figure}
\centering
\includegraphics[width=\textwidth]{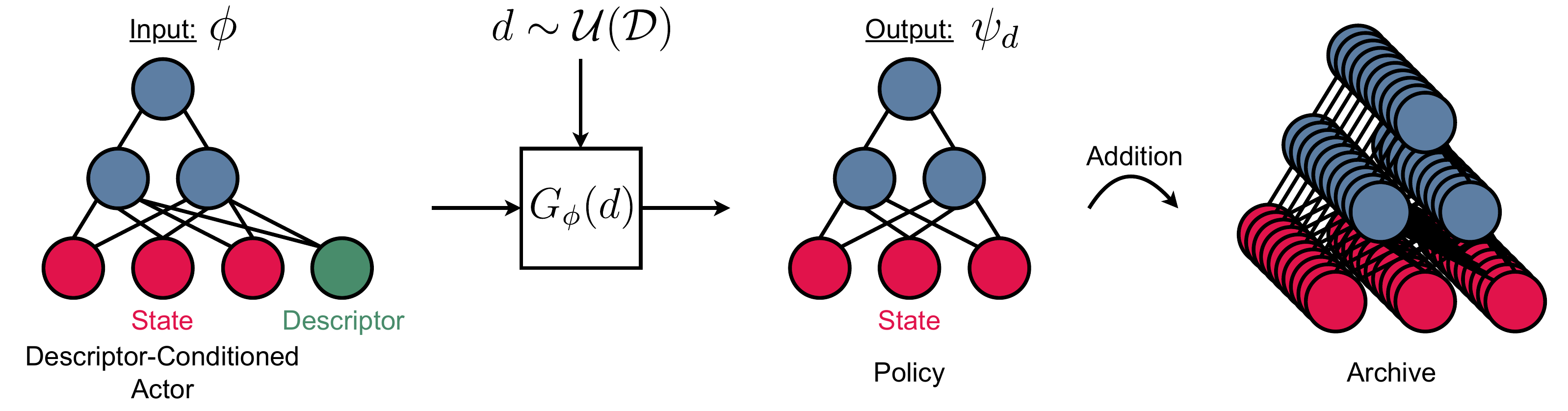}
\caption{\textbf{Descriptor-Conditioned Actor Injection} transforms the generally capable descriptor-conditioned actor $\policy_\actorParams(\obs \mid \desc)$ into a specialized, unconditioned policy $\policy_{\policyParams_d}(\obs)$. For a given descriptor $d$, it generates a policy $G_\actorParams(d) = \policy_{\policyParams_d}$ with an architecture matching the policies in the archive. This enables the injection of specialized versions of the actor into different niches of the solution space.}
\label{fig:actor-injection}
\Description{Descriptor-Conditioned Actor Injection.}
\end{figure}

The descriptor-conditioned actor is a function $\policy_\actorParams \colon \obsSpace \times \descSpace \to \actionSpace$. This means it takes as input both a state from the observation space $\obsSpace$ and a descriptor from the descriptor space $\descSpace$, and outputs an action from the action space $\actionSpace$.
We can define a generative model $G_\actorParams: \descSpace \to \policySpace$ that maps from the descriptor space to the policy space.
For any descriptor $d \in \descSpace$, we have: $G_\actorParams(d) = \policy_{\policyParams_d}$, where $\policy_{\policyParams_d}$ is a policy in the policy space $\policySpace$ that is specialized for the descriptor $d$. This specialized policy can be defined as: $\policy_{\policyParams_d}(s) = \policy_\actorParams(\obs \mid \desc)$ for all $s \in \obsSpace$.
In other words, $G_\actorParams(d)$ is a policy that, for any given state $s$, produces the same action as the descriptor-conditioned actor would produce for that state and the fixed descriptor $d$.
This generative model allows us to create specialized policies for any given descriptor, effectively turning our descriptor-conditioned actor into a generator of policies tailored to specific descriptors.

However, the produced parameters $\policyParams_d$ need to have the exact same architecture as the policies in the population.
The actor and the policies in the archive share a common architecture of $n$ layers of $h$ hidden units each, except one crucial difference: the first layer of the actor is designed to accept both state and descriptor inputs, while the policies are structured to receive only state inputs, see \cref{fig:actor-injection}.
Mathematically, let $\weight$ and $\bias$ denote the weights and biases of the first layer of the descriptor-conditioned actor, respectively. For any state $\obs$ and descriptor $\desc$, the computation of the first layer can be expressed as:
$$(\obs || \desc)^\intercal \weight + \bias = \obs^\intercal \weight_1 + (\desc^\intercal \weight_2 + \bias)$$
where $\weight_1$ is a matrix of dimension $(\dim(\obsSpace), h)$ and $\weight_2$ is a matrix of dimension $(\dim(\descSpace), h)$. This formulation allows us to reinterpret the computation as the state $\obs$ multiplied with the matrix $\weight_1$ plus a modified bias term $d^\intercal \weight_2 + \bias$. Notably, $\weight_1$ and the modified bias $d^\intercal \weight_2 + \bias$ have the same dimensions as the corresponding components in the archive policies. Thus, as the remaining layers have the same size, we can recompute the parameters of the first layer, in order to exactly match the architectures and inject the specialized versions of the descriptor-conditioned actor in the archive.

In our implementation, we sample $\aibatchsize = 64$ descriptors at each generation $\desc_1, ..., \desc_{\aibatchsize}$ in the descriptor space $\descSpace$. For each sampled descriptor, we create a specialized version of the actor by recomputing its parameters as described above. These specialized policies are then proposed for addition to the archive, effectively injecting multiple variations of our descriptor-conditioned actor into the population, see \MyRef{alg:dcrl-me-ai}.

The Actor Injection mechanism in \dcrlme{} not only enhances the quality and diversity of the archive but also elegantly addresses the actor evaluation challenges faced by \dcgme{}. In \dcgme{}, separate actor evaluations were necessary to generate transitions with both target and observed descriptors, crucial for training the descriptor-conditioned actor-critic model, as explained in \cref{sec:introduction}. This process, while effective, reduced sample efficiency. In contrast, \dcrlme{}'s Actor Injection approach inherently solves this issue. By using the descriptor-conditioned actor as a generative model to produce policies for specific descriptors, each injected policy naturally encapsulates both the target descriptor (used for generation) and the observed descriptor (resulting from policy execution). This dual-descriptor information is intrinsically present in the transitions generated by these injected policies during evaluation. Consequently, \dcrlme{} eliminates the need for separate actor evaluations, significantly improving sample efficiency. The algorithm now obtains the required descriptor-rich transitions directly from the evaluation of injected policies, which serve the dual purpose of populating the archive and providing training data for the actor-critic model. This elegant solution not only resolves the mismatch between unconditioned archive policies and the descriptor-conditioned learning algorithm but also streamlines the overall process, making \dcrlme{} more efficient and effective in its exploration of the solution space.

\section{Experiments}
\label{sec:experiments}
Each experiment is replicated 20 times with random seeds, over one million evaluations and the implementations are based on the QDax library~\citep{chalumeau2023qdax}.
The full source code is available at \href{https://github.com/adaptive-intelligent-robotics/DCRL-MAP-Elites}{github.com/adaptive-intelligent-robotics/DCRL-MAP-Elites}, in a containerized environment in which all the experiments and figures can be reproduced.
For the quantitative results, we report p-values based on the Wilcoxon–Mann–Whitney $U$ test with Holm-Bonferroni correction.

\subsection{Tasks}
We evaluate \dcrlme{} and \dcgme{} on seven continuous control locomotion QD tasks~\citep{nilsson_PolicyGradientAssisted_2021} implemented in Brax~\citep{brax} and derived from standard RL benchmarks, see \cref{tab:tasks}. Ant Omni, AntTrap Omni and Humanoid Omni are \emph{omnidirectional} tasks, in which the objective is to minimize energy consumption and the descriptor is the final position of the agent. Walker Uni, HalfCheetah Uni, Ant Uni and Humanoid Uni are \emph{unidirectional} task in which the objective is  to go forward as fast as possible while minimizing energy consumption and the descriptor is the feet contact rate for each foot of the agent.
Walker Uni, HalfCheetah Uni, Ant Uni were introduced in \pgame{} paper~\citep{nilsson_PolicyGradientAssisted_2021} and Humanoid Uni, Ant Omni, Humanoid Omni were introduced by~\citet{flageat_BenchmarkingQualityDiversityAlgorithms_2022}.
AntTrap Omni is adapted from \qdpg{} paper~\citep{pierrot_DiversityPolicyGradient_2022}, the only difference being the elimination of the forward term in the reward function. We introduce AntTrap Omni to evaluate \dcrlme{} and \dcgme{} on a deceptive, omnidirectional environment. The trap creates a discontinuity of fitness in the descriptor space as points on both sides of the trap are close, but require two different trajectories to achieve these descriptors. Thus, the descriptor-conditioned critic needs to learn that discontinuity to provide accurate policy gradients.

\pgame{} has previously shown state-of-the-art results on unidirectional tasks, in particular Walker Uni, HalfCheetah Uni and Ant Uni, but tends to struggle on omnidirectional tasks. In omnidirectional tasks, the global maximum of the fitness function is a solution that does not move, which is directly opposed to discovering how to reach different locations. Hence, the offsprings generated by the PG variation operator will tend to move less and travel a shorter distance. Instead, \dcgme{} aims to improve the energy consumption while maintaining the ability to reach distant locations.

\begin{table}[h]
\caption{Evaluation Tasks}
\label{tab:tasks}
\centering
\newdimen\length
\length=1.cm
\tabcolsep=0.1cm
\begin{tabular}{l | c c c c c c c}
    \toprule
    & \textsc{Ant} & \textsc{AntTrap} &
    \textsc{Humanoid} & \textsc{Walker} & \textsc{HalfCheetah} & \textsc{Ant} & \textsc{Humanoid}\\
    & \textsc{Omni} & \textsc{Omni} &
    \textsc{Omni} & \textsc{Uni} & \textsc{Uni} & \textsc{Uni} & \textsc{Uni}\\
    & \includegraphics[height=0.8\length, width=\length]{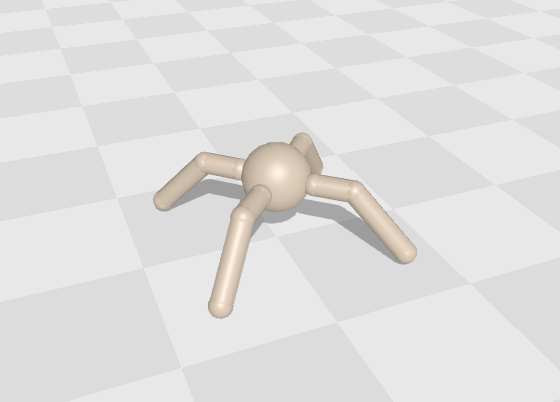} & \includegraphics[height=0.8\length, width=\length]{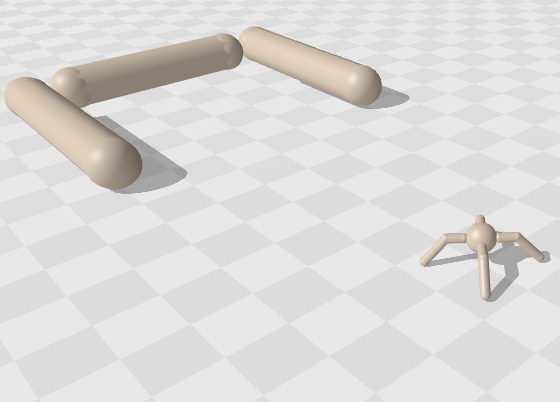} & \includegraphics[height=0.8\length, width=\length]{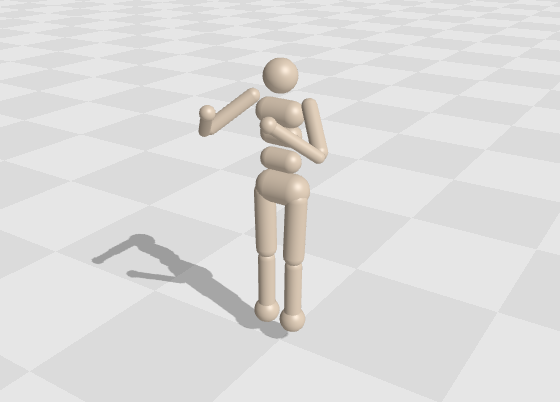} & \includegraphics[height=0.8\length, width=\length]{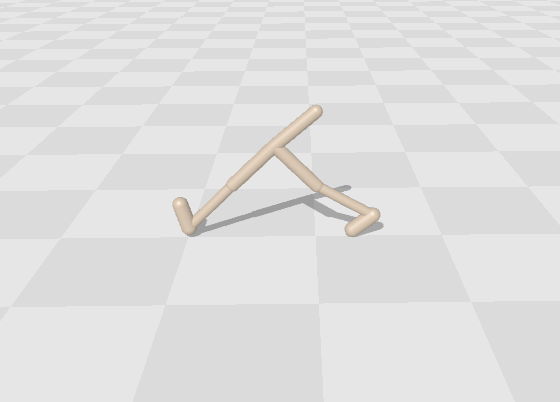} & \includegraphics[height=0.8\length, width=\length]{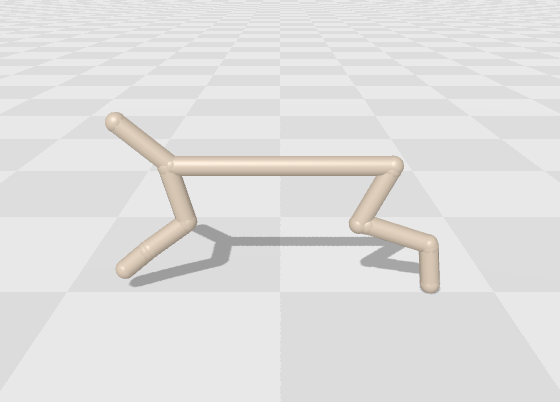} & \includegraphics[height=0.8\length, width=\length]{images/ant.png} &
    \includegraphics[height=0.8\length, width=\length]{images/humanoid.png}\\
    \midrule
    \textsc{State} & \multicolumn{7}{c}{Position and velocity of body parts}\\
    \textsc{Action} & \multicolumn{7}{c}{Torques applied at the hinge joints}\\
    \textsc{State dim} & 30 & 30 & 245 & 18 & 19 & 28 & 245\\
    \textsc{Action dim} & 8 & 8 & 17 & 6 & 6 & 8 & 17\\
    \textsc{Descriptor dim} & 2 & 2 & 2 & 2 & 2 & 4 & 2\\
    \textsc{Episode len} & 250 & 250 & 1000 & 1000 & 1000 & 1000 & 1000\\
    \textsc{Parameters} & 21,512 & 21,512 & 50,193 & 19,718 & 19,846 & 21,512 & 50,193\\
    \bottomrule
\end{tabular}
\end{table}

\subsection{Main Results}
\subsubsection{Baselines}
We compare \dcrlme{} and \dcgme{}~\cite{faldor_dcgme_2023} with four algorithms, namely \me~\citep{vassiliades_UsingCentroidalVoronoi_2017}, \mees{}~\citep{colas_ScalingMAPElitesDeep_2020}, \pgame{}~\citep{nilsson_PolicyGradientAssisted_2021}, and \qdpg{}~\citep{pierrot_DiversityPolicyGradient_2022}.

\subsubsection{Metrics}
\label{sec:archive-metrics}
We consider the QD score, coverage and max fitness to evaluate the final populations (i.e. archives) of all algorithms throughout training, as defined in \citet{pugh_QualityDiversityNew_2016,flageat_BenchmarkingQualityDiversityAlgorithms_2022} and used in \pgame{} paper~\citep{nilsson_PolicyGradientAssisted_2021}.
The main metric is the \emph{QD score}, which represents the sum of fitness of all solutions stored in the archive. This metric captures both the quality and the diversity of the population. In the tasks considered, the fitness is always positive, which avoids penalizing algorithms for finding additional solutions.
We also consider the \emph{coverage}, which represents the proportion of filled cells in the archive, measuring descriptor space illumination.
Finally, we also report the \emph{max fitness}, which is defined as the fitness of the best solution in the archive.

\subsubsection{Results}
\begin{figure}
\includegraphics[width=\textwidth]{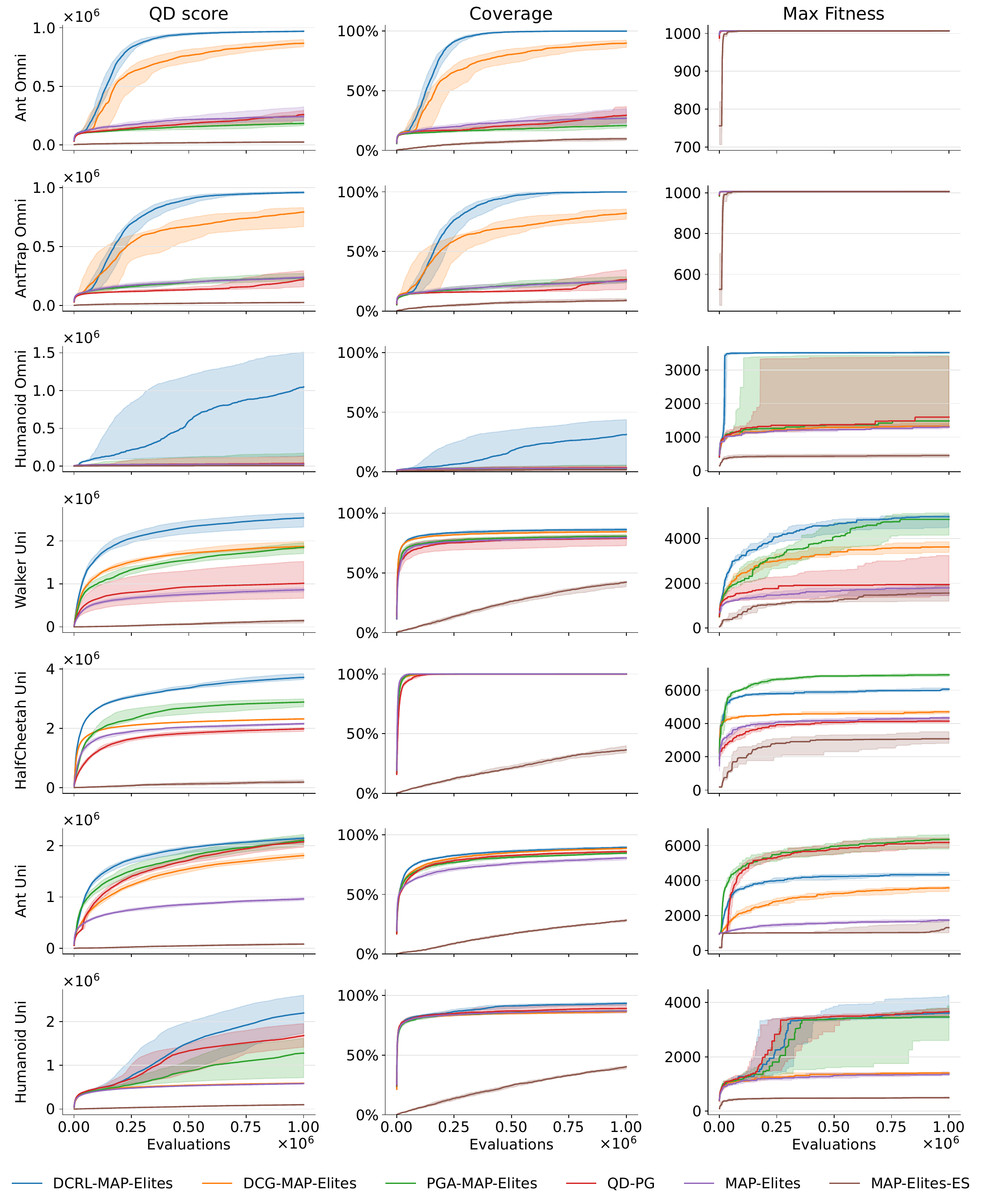}
\caption{QD score, coverage and max fitness for \dcrlme{}, \dcgme{} and all baselines on all tasks. Each experiment is replicated 20 times with random seeds. The solid line is the median and the shaded area represents the first and third quartiles.}
\label{fig:plot-main}
\Description[\dcgme{} outperforms all baselines in QD score, coverage and max fitness on most tasks.]{\dcgme{} outperforms all baselines, on all metrics and on all tasks, except on HalfCheetah Uni and on Ant Uni}
\end{figure}
The experimental results presented in \cref{fig:plot-main} demonstrate that \dcrlme{} achieves equal or higher QD score and coverage than all baselines on all tasks, especially \dcgme{} and \pgame{}, the previous state-of-the-art. On Ant Uni and Humanoid Uni, \dcgme{} achieves a higher median QD score but not significantly. On all other tasks, \dcgme{} achieves a significantly higher QD score ($p < 0.003$), demonstrating that our method generates populations of solutions that are higher-performing and more diverse.
Especially, the coverage metric shows that \dcrlme{} and \dcgme{} surpass the exploration capabilities of \qdpg{} on all tasks ($p < 0.05$).
\dcrlme{} significantly outperforms \dcgme{}~\citep{faldor_dcgme_2023} on all environments except Ant Uni ($p < 0.01)$, where they perform similarly, showing that the improvements made to the algorithm are beneficial.
\dcrlme{} also achieves equal or significantly better max fitness on all environments except on HalfCheetah Uni and Ant Uni, where \pgame{} is better, showing room for improvement.
Finally, we also show that our method still benefits from the exploration power of the GA operator even in deceptive environment like AntTrap Omni.
The experimental results confirm that \dcrlme{} and \dcgme{} is able to overcome the limits of \pgame{} on omnidirectional tasks while performing better on the unidirectional tasks ($p < 0.005$) except Ant Uni where our method is not significantly better.
Thus, confirming the interest of having a descriptor-conditioned gradient to make the PG variation operator fruitful in a wider range of tasks.
Overall, \dcrlme{} and \dcgme{} show competitive performance on all metrics and tasks, hence proving to be the first successful effort in the QD-RL literature to achieve well on both the unidirectional and omnidirectional tasks. Previous efforts were usually adapted to either one or the other~\citep{nilsson_PolicyGradientAssisted_2021,pierrot_DiversityPolicyGradient_2022,tjanaka_ApproximatingGradientsDifferentiable_2022}.
Qualitative results in \cref{fig:plot-archive} also show that \dcrlme{} discovers solutions that are more diverse and higher-performing than other baselines on Ant Omni task. The final archives for all algorithms and on all tasks are provided in \cref{appendix:archives}.
\begin{figure}[H]
\centering
\includegraphics[width=\textwidth]{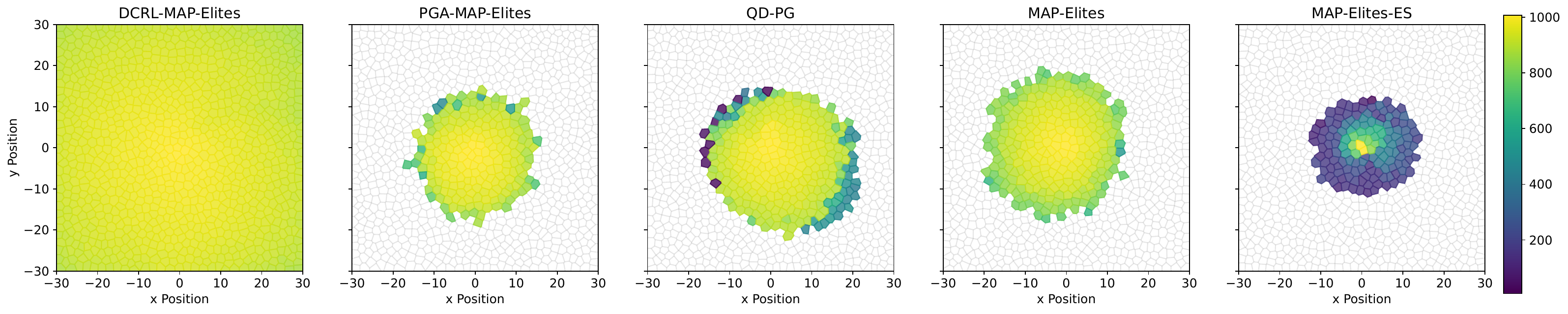}
\caption{\textbf{Ant Omni} Archive at the end of training for all algorithms.}
\label{fig:plot-archive}
\Description[\dcgme{}'s archive significantly outperforms all baselines in terms of quality and diversity]{\dcgme{}'s archive significantly outperforms all baselines in terms of quality and diversity}
\end{figure}

\subsection{Ablations}
\subsubsection{Ablation studies}
To better understand the contributions of different components in our approach, we conducted ablation studies comparing \dcrlme{} and \dcgme{} with two variants. The first, Ablation AI, lacks both actor injection and actor evaluation, eliminating the provision of descriptor-conditioned active samples to the \tdthree{} algorithm. The second, Ablation Actor, features a non-descriptor-conditioned actor, removing the archive distillation component while maintaining a descriptor-conditioned critic. For reference, \dcgme{} does not include actor injection but performs actor evaluation to provide descriptor-conditioned active samples to the \tdthree{} algorithm.

\subsubsection{Results}
\begin{figure}
\includegraphics[width=\textwidth]{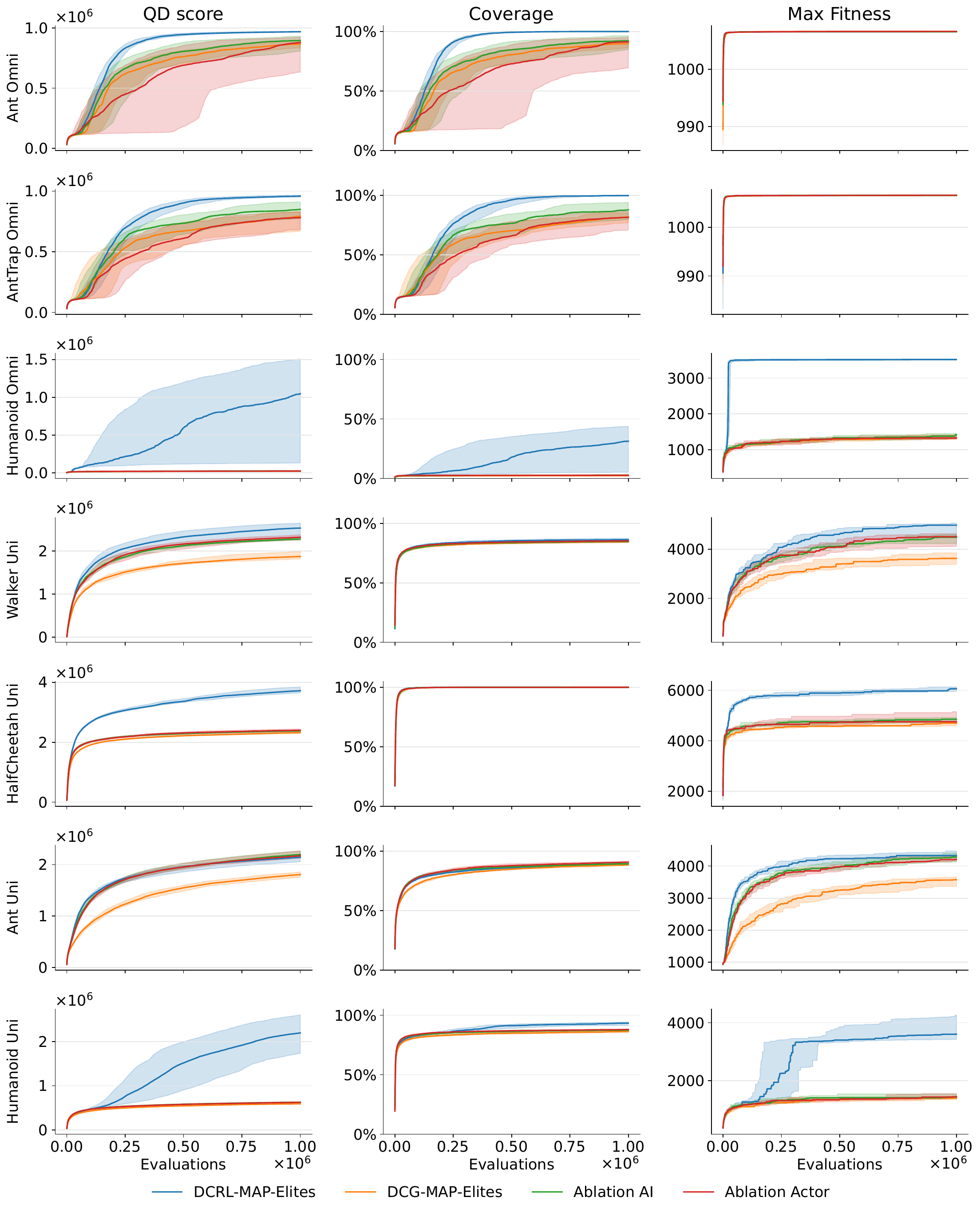}
\caption{QD score, coverage and max fitness for \dcrlme{}, \dcgme{} and the ablations on all tasks. Each experiment is replicated 20 times with random seeds. The solid line is the median and the shaded area represents the first and third quartiles.}
\label{fig:plot-ablation}
\Description[\dcgme{} outperforms all ablations, on all metrics and on all tasks.]{\dcgme{} outperforms all ablations, on all metrics and on all tasks.}
\end{figure}
The results presented in Figure \ref{fig:plot-ablation} underscore the critical role of each component in DCRL-MAP-Elites, with the full implementation consistently outperforming ablated versions across various performance metrics and tasks.
Actor injection proves significantly beneficial in terms of QD score, on all tasks ($p < 0.05$) except Ant Uni where they perform comparably.
Having a descriptor-conditioned actor $\policy_\actorParams(\,. \mid d)$ rather than a normal actor $\policy_\actorParams(\,. \,)$ proves significantly beneficial in terms of QD score, on all tasks ($p < 10^{-4}$), demonstrating that the descriptor-conditioned actor enables archive distillation while being beneficial for the critic's training.
\dcgme{} achieves equal or higher QD score than the AI ablation, showing the importance of on-policy samples.
Overall, \dcrlme{} demonstrate higher or comparable performance on all metrics and all tasks compared to \dcgme{} and to the ablations, hence proving the importance of the different enhancements compared to \pgame{}.

\subsection{Reproducibility}
\subsubsection{Reproducibility Metrics}
\label{sec:reproducibility-metrics}
We also consider three metrics to evaluate the reproducibility of the final archives for all algorithms and of the descriptor-conditioned actor for \dcrlme{}, at the end of training.
QD algorithms based on \me{} output a population of solutions that we evaluate with the QD score, coverage and max fitness, see \cref{sec:archive-metrics}. However, these metrics can be misleading because in stochastic environments, a solution might give different fitnesses and descriptors when evaluated multiple times. Consequently, the QD score, coverage and max fitness can be overestimated, an effect that is well-known and that has been studied in the past~\citep{flageat2023uncertain}.
An archive of solutions is considered reproducible, if the QD score, coverage and max fitness does not change substantially after multiple reevaluation of the individuals. Thus, to assess the reproducibility of the archives, we consider the \emph{expected QD score}, the \emph{expected distance to descriptor} and the \emph{expected max fitness}.
To calculate those metrics, we reevaluate each solution in the archive 512 times, to approximate its expected fitness and expected distance to descriptor. The expected distance to descriptor of a solution is simply the expected euclidean distance between the descriptor of the cell of the individual and the observed descriptors. Therefore, for the expected distance to descriptor, lower is better.
We use the expected fitness and expected distance to descriptor of all solutions to calculate the expected QD score, expected distance to descriptor and expected max fitness of the archive.

Additionally, \dcrlme{}'s descriptor-conditioned actor can in principle achieve different descriptors and thus, is comparable to an archive. Similarly to the archive, we evaluate its expected QD score, expected distance to descriptor and expected max fitness. To that end, we take the descriptor $d$ of each filled cell in the corresponding archive, and evaluate the actor $\policy_\actorParams(\, . \mid d)$ 512 times, to approximate its expected fitness and expected distance to descriptor.
Analogously to the archive, we use the expected fitnesses and expected distances to descriptor to calculate the expected QD score, expected distance to descriptor and expected max fitness of the descriptor-conditioned actor.

\subsubsection{Results}
\begin{figure}
\includegraphics[width=\textwidth]{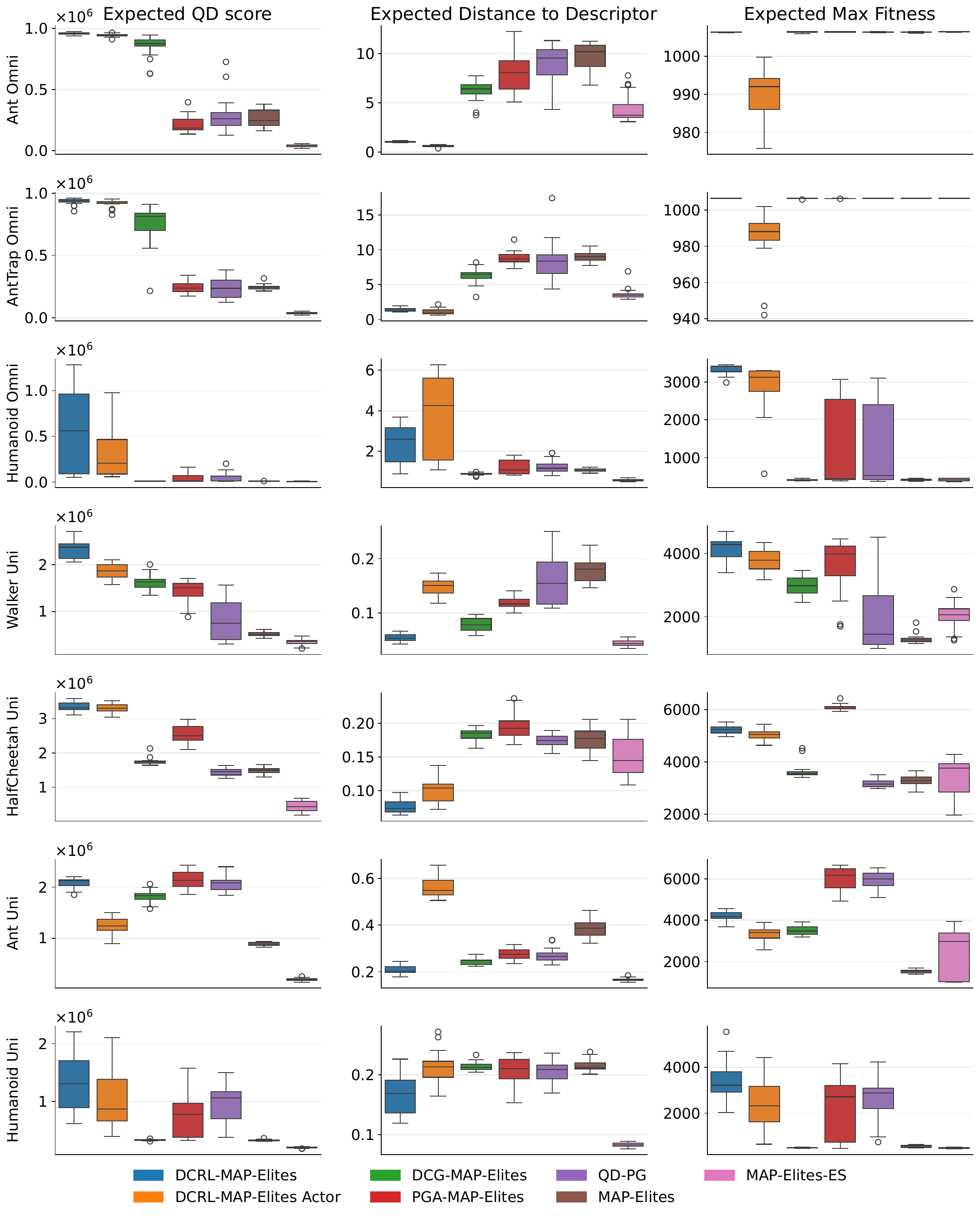}
\caption{Expected QD score, expected distance to descriptor (lower is better) and expected max fitness for \dcrlme{}, the descriptor-conditioned policy and the baselines on all tasks. Each experiment is replicated 20 times with random seeds.}
\label{fig:reproducibility}
\Description[\dcgme{} outperforms all baselines, on all reproducibility metrics and on all tasks.]{\dcgme{} outperforms all baselines, on all reproducibility metrics and on all tasks.}
\end{figure}
In \cref{fig:reproducibility}, we provide the expected QD score, expected distance to descriptor and expected max fitness of the final archive and the descriptor-conditioned policy, see \cref{sec:reproducibility-metrics}.
First, we can see that \dcrlme{}'s final archive achieves equal or higher expected QD score than all baselines on all tasks. The descriptor-conditioned actor performs similarly to \dcrlme{}'s archive on most environments, but performs significantly worse on Ant Uni. This shows that, in most cases, the descriptor-conditioned actor is able to restore the quality of the archive although having compressed the information in a single network.
Second, \dcrlme{} obtains better expected distance to descriptor (lower is better) than all baselines except \mees{} on all tasks. However, \mees{} obtains worse QD score and most importantly, worst coverage, making it easier for \mees{} to achieve a low expected distance to descriptor. \dcrlme{}'s actor obtains similar expected distance to descriptor on omnidirectional. However, it performs consistently slightly worse on unidirectional tasks. This shows that in some cases, while compressing the quality of the archive in a single network, the descriptor-conditioned actor can also exhibit the same diversity as the population.
Those two combined observations show that the final archive and descriptor-conditioned policy have similar properties on omnidirectional tasks. Overall, those results show that our single generally capable actor can already be seen as a promising summary of our archive, showing very similar properties on most tasks.

Finally, \dcrlme{} performs significantly better than \dcgme{} on all tasks and all robustness metrics, except on Humanoid Omni, where \dcgme{} obtains a slightly better expected distance to descriptor. However, on this task, \dcgme{} obtains significantly worse coverage than \dcrlme{} (see \cref{fig:plot-main}), which makes it easier for \dcgme{} to achieve better reproducibility results. Overall, the reproduciblity results are in line with the main results shown in \cref{fig:plot-main}: \dcrlme{} achieves better expected QD score than all other baselines on all tasks.

\subsection{Variation Operators Evaluation}
\subsubsection{Variation Operator Metrics}
\label{sec:variation-operator-metrics}
This section presents an empirical analysis of the variation operators in \pgame{} and \dcrlme{}, aiming to elucidate their synergies and relative contributions to performance improvement. Our objectives are twofold: (1) evaluating the contribution of each variation operator, (2) assessing the synergies between the different variation operators.

Both \dcrlme{} and \pgame{} employ the same GA variation operator, producing 128 offspring per generation. \pgame{} additionally uses a PG variation operator for another 128 offspring, while \dcrlme{} utilizes a descriptor-conditioned PG variation operator for 64 offspring and an actor injection mechanism for 64 offspring, maintaining a total batch size of 256 offspring in both cases.

To evaluate operator performance, we introduce the metric of \emph{cumulative improvement}. This metric measures the accumulated fitness improvement of offspring added to the archive by each variation operator throughout training. By tracking cumulative improvement, we can analyze the interaction and dynamics between the different variation operators and actor injection, providing insights into their relative contributions and effectiveness over time.

\subsubsection{Results}
\begin{figure}
\includegraphics[width=\textwidth]{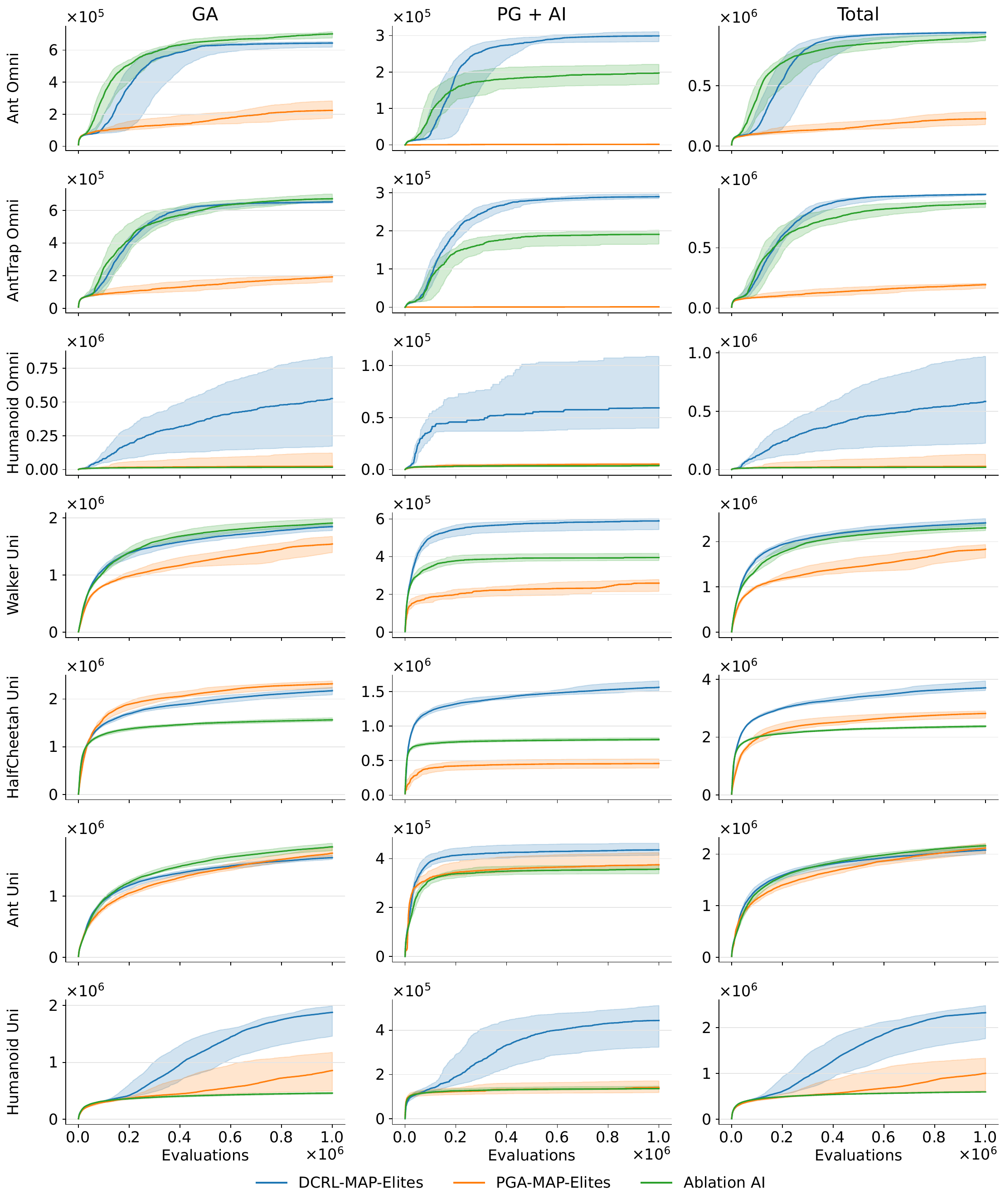}
\caption{Cumulative improvement (\cref{sec:variation-operator-metrics}) for (\textbf{top}) GA variation operator and (\textbf{bottom}) PG variation operator plus Actor Injection (AI). Each experiment is replicated 20 times with random seeds. The solid line is the median and the shaded area represents the first and third quartiles.}
\label{fig:improvement}
\Description[Accumulated number of offsprings added to the archive for \dcgme{}, \pgame{}, and Ablation AI, for each variation operator. \dcgme{} achieves a higher number of elites on all tasks.]{Accumulated number of offsprings added to the archive for \dcgme{}, \pgame{}, and Ablation AI, for each variation operator. \dcgme{} achieves a higher number of elites on all tasks.}
\end{figure}
Figure \ref{fig:improvement} presents the cumulative improvement for the GA variation operator (top row) and the PG + AI variation operator (bottom row) for \dcrlme{}, \pgame{}, and Ablation AI throughout training. Each line corresponds to an emitter with a batch size of 128.

The results reveal several key insights into the performance of the variation operators.
First, the effectiveness of descriptor-conditioning is demonstrated by the performance of Ablation AI, which generates greater cumulative improvement than \pgame{}. This demonstrates that the descriptor-conditioned critic alone produces higher-performing solutions than the traditional critic in \pgame{}, highlighting the value of incorporating descriptor information into the critic's learning process.

Second, the GA operator in Ablation AI and \dcrlme{} consistently achieves higher cumulative improvement compared to its counterpart in \pgame{}, despite being identical. This suggests a synergy between operators, where the solutions generated by the descriptor-conditioned PG operator and actor injection in \dcrlme{} serve as superior stepping stones, enhancing the GA operator's effectiveness in finding improved solutions.

Third, \dcrlme{} with actor injection achieves even higher cumulative improvement than Ablation AI, underscoring the complementary nature of the descriptor-conditioned PG operator and actor injection in generating diverse, high-quality solutions.

Finally, task-specific performance variations are also observed. In complex environments like Humanoid, the actor injection mechanism in \dcrlme{} appears to enable navigating the solution space more effectively, resulting in higher cumulative improvement compared to \pgame{} and Ablation AI. In other words, actor injection seems crucial in solving Humanoid Omni and Uni tasks.

These findings demonstrate the complex interplay between variation operators in \dcrlme{}, showcasing how the descriptor-conditioned approach and actor injection mechanism synergize to enhance overall performance. The cumulative improvement metric provides a nuanced view of each operator's contribution, revealing not just the quantity of archive additions but the quality of improvements over time.

In conclusion, this analysis underscores the importance of carefully designed operator combinations in QD algorithms. The superior performance of \dcrlme{} across various tasks highlights the potential of descriptor-conditioned approaches and strategic actor injection in enhancing both the quality and diversity of discovered solutions.

\section{Related Work}
\label{sec:related-work}
\subsection{Scaling QD to Neuroevolution}
The challenge of evolving diverse solutions in a high-dimensional search space has been an active research subject over recent years.
\mees{}~\citep{colas_ScalingMAPElitesDeep_2020} scales \me{} to high-dimensional solutions parameterized by large neural networks. This algorithm leverages Evolution Strategies (ES) to perform a directed search that is more efficient than random mutations used in Genetic Algorithms~\citep{salimans2017evolution}. Fitness and novelty gradients are estimated locally from many perturbed versions of the parent solution to generate a new one. The population tends towards regions of the parameter space with higher fitness or novelty but it requires to sample and evaluate a large number of solutions, making it particularly data inefficient.
To improve sample efficiency, methods that combine \me{} with RL~\citep{nilsson_PolicyGradientAssisted_2021,pierrot_DiversityPolicyGradient_2022, tjanaka_ApproximatingGradientsDifferentiable_2022,pierrot_EvolvingPopulationsDiverse_2023} have emerged and use time step level information to efficiently evolve populations of high-performing and diverse neural network for complex tasks. 
\pgame{}~\citep{nilsson_PolicyGradientAssisted_2021} uses policy gradients for part of its mutations, see \cref{sec:background-pga-me} for details.
\cmamega{}~\citep{tjanaka_ApproximatingGradientsDifferentiable_2022} estimates descriptor gradients with ES and combines the fitness gradient and the descriptor gradients with a \cmaes{} mechanism~\citep{hansen_CMAEvolutionStrategy_2023,fontaine_DifferentiableQualityDiversity_2021}.
\qdpg{}~\citep{pierrot_DiversityPolicyGradient_2022} introduces a diversity reward based on the novelty of the states visited and derives a policy gradient for the maximization of those diversity rewards which helps exploration in settings where the reward is sparse or deceptive.
\pbtme{}~\citep{pierrot_EvolvingPopulationsDiverse_2023} mixes \me{} with a population based training process~\citep{jaderberg_PopulationBasedTraining_2017} to optimize hyper-parameters of diverse RL agents.
Interestingly, recent work~\citep{tjanaka_TrainingDiverseHighDimensional_2023} scales the algorithm \cmamae{}~\citep{fontaine_CovarianceMatrixAdaptation_2023} to high-dimensional policies on robotics tasks with pure ES while showing comparable data efficiency to QD-RL approaches, but is still outperformed by \pgame{}.

\subsection{Conditioning the critic}
None of the methods described in the previous section use a descriptor-conditioned critic nor descriptor-conditioned policies as our method does.
The concept of descriptor-conditioned critic is related to Universal Value Function Approximators~\citep{schaul_UniversalValueFunction_2015}, extensively used in skill discovery reinforcement learning, a field that share a similar motivation to QD~\citep{chalumeau_NeuroevolutionCompetitiveAlternative_2022}. In \vic{}, \diayn{}, \dads{}, \smerl{}~\citep{gregor_VariationalIntrinsicControl_2016,eysenbach_DiversityAllYou_2018,sharma_DynamicsAwareUnsupervisedDiscovery_2019,kumar_OneSolutionNot_2020}, the actors and critics are conditioned on a sampled prior but does not correspond to a real posterior like in \dcrlme{}. Furthermore, those methods use a notion of diversity defined at the step-level rather than trajectory-level like \dcrlme{}. Moreover, they do not use an archive to store a population, resulting in much smaller sets of final policies. Finally, it has been shown that QD methods are competitive with skill discovery reinforcement learning algorithms~\citep{chalumeau_NeuroevolutionCompetitiveAlternative_2022}, specifically for adaptation and hierarchical learning.

\subsection{Archive distillation}
Distilling the knowledge of an archive into a single policy is an alluring process that reduces the number of parameters outputted by the algorithm and enables generalization and interpolation/extrapolation. Although distillation is usually referring to policy distillation --- learning the observation/action mapping from a teacher policy --- we present archive distillation as a general term referring to any kind of knowledge transfer from an archive to another model, should it be the policies, transitions experienced in the environment, full trajectories or discovered descriptors.

To the best of our knowledge, only two QD-related works use the concept of archive distillation. Go-Explore~\citep{ecoffet_FirstReturnThen_2021} keeps an archive of states and trains a goal-conditioned policy to reproduce the trajectory of the policy that reached that state. Another related approach is to learn a generative policy network~\citep{jegorova_BehaviouralRepertoireGenerative_2019} over the policies contained in the archive. Our approach \dcrlme{} distills the experience of the archive into a single versatile policy.

\section{Conclusion}
\label{sec:conclusion}
In this work, we introduce \dcrlme{}, an extension of \dcgme{} that fully leverages both the actor and critic of the RL algorithm. While \dcgme{} primarily utilized the critic, \dcrlme{} also harnesses the descriptor-conditioned actor as a generative model to produce diverse policies, which are then injected into the offspring batch at each generation.
Similar to \dcgme{}, our method leverages a descriptor-conditioned actor-critic model that enhances the PG variation operator, while simultaneously distilling the archive into a single policy.
Our method, \dcrlme{}, achieves equal or better performance than all baselines on seven continuous control locomotion tasks. Additionally, we complement the main result with an ablation study and two empirical analyses. First, we demonstrate that \dcrlme{} and \dcgme{} achieve better reproduciblity in the face of environment stochasticity. Second, we also show that the descriptor-conditioned PG variation operator synergizes with the GA variation operator to enhance the overall performance.
The superior performance of \dcrlme{} across various tasks highlights the potential of descriptor-conditioned approaches and strategic actor injection in enhancing both the quality and diversity of discovered solutions.

The benefits of combining RL methods with \pgame{} come with the limitations of MDP settings. Specifically, we are limited to evolving differentiable solutions and the foundations of RL algorithms rely on the Markov property and full observability. In this work in particular, we face challenges with the Markov property because the descriptors depend on full trajectories. Thus, the scaled reward introduced in our method depends on the full trajectory and not only on the current state and action. The performance of the descriptor-conditioned policy also shows that there is room for improvement to better distill the knowledge of the archive.
For future work, we would like to leverage the descriptor-conditioned critic to mutate solutions to produce offspring towards different descriptors, thereby explicitly promoting diversity.

\endgroup

\bibliographystyle{ACM-Reference-Format}
\bibliography{bibliography}


\begin{thebibliography}{48}


\ifx \showCODEN    \undefined \def \showCODEN     #1{\unskip}     \fi
\ifx \showDOI      \undefined \def \showDOI       #1{#1}\fi
\ifx \showISBNx    \undefined \def \showISBNx     #1{\unskip}     \fi
\ifx \showISBNxiii \undefined \def \showISBNxiii  #1{\unskip}     \fi
\ifx \showISSN     \undefined \def \showISSN      #1{\unskip}     \fi
\ifx \showLCCN     \undefined \def \showLCCN      #1{\unskip}     \fi
\ifx \shownote     \undefined \def \shownote      #1{#1}          \fi
\ifx \showarticletitle \undefined \def \showarticletitle #1{#1}   \fi
\ifx \showURL      \undefined \def \showURL       {\relax}        \fi
\providecommand\bibfield[2]{#2}
\providecommand\bibinfo[2]{#2}
\providecommand\natexlab[1]{#1}
\providecommand\showeprint[2][]{arXiv:#2}

\bibitem[Andrychowicz et~al\mbox{.}(2017)]%
        {andrychowicz_HindsightExperienceReplay_2017}
\bibfield{author}{\bibinfo{person}{Marcin Andrychowicz}, \bibinfo{person}{Filip Wolski}, \bibinfo{person}{Alex Ray}, \bibinfo{person}{Jonas Schneider}, \bibinfo{person}{Rachel Fong}, \bibinfo{person}{Peter Welinder}, \bibinfo{person}{Bob McGrew}, \bibinfo{person}{Josh Tobin}, \bibinfo{person}{OpenAI Pieter~Abbeel}, {and} \bibinfo{person}{Wojciech Zaremba}.} \bibinfo{year}{2017}\natexlab{}.
\newblock \showarticletitle{Hindsight {Experience} {Replay}}. In \bibinfo{booktitle}{\emph{Advances in {Neural} {Information} {Processing} {Systems}}}, Vol.~\bibinfo{volume}{30}. \bibinfo{publisher}{Curran Associates, Inc.}
\newblock
\urldef\tempurl%
\url{https://proceedings.neurips.cc/paper_files/paper/2017/hash/453fadbd8a1a3af50a9df4df899537b5-Abstract.html}
\showURL{%
\tempurl}


\bibitem[Chalumeau et~al\mbox{.}(2022)]%
        {chalumeau_NeuroevolutionCompetitiveAlternative_2022}
\bibfield{author}{\bibinfo{person}{Felix Chalumeau}, \bibinfo{person}{Raphael Boige}, \bibinfo{person}{Bryan Lim}, \bibinfo{person}{Valentin Macé}, \bibinfo{person}{Maxime Allard}, \bibinfo{person}{Arthur Flajolet}, \bibinfo{person}{Antoine Cully}, {and} \bibinfo{person}{Thomas Pierrot}.} \bibinfo{year}{2022}\natexlab{}.
\newblock \showarticletitle{Neuroevolution is a {Competitive} {Alternative} to {Reinforcement} {Learning} for {Skill} {Discovery}}.
\newblock
\urldef\tempurl%
\url{https://openreview.net/forum?id=6BHlZgyPOZY}
\showURL{%
\tempurl}


\bibitem[Chalumeau et~al\mbox{.}(2023)]%
        {chalumeau2023qdax}
\bibfield{author}{\bibinfo{person}{Felix Chalumeau}, \bibinfo{person}{Bryan Lim}, \bibinfo{person}{Raphael Boige}, \bibinfo{person}{Maxime Allard}, \bibinfo{person}{Luca Grillotti}, \bibinfo{person}{Manon Flageat}, \bibinfo{person}{Valentin Macé}, \bibinfo{person}{Arthur Flajolet}, \bibinfo{person}{Thomas Pierrot}, {and} \bibinfo{person}{Antoine Cully}.} \bibinfo{year}{2023}\natexlab{}.
\newblock \bibinfo{title}{QDax: A Library for Quality-Diversity and Population-based Algorithms with Hardware Acceleration}.
\newblock
\newblock
\showeprint[arxiv]{2308.03665}~[cs.AI]


\bibitem[Chatzilygeroudis et~al\mbox{.}(2021)]%
        {chatzilygeroudis_QualityDiversityOptimizationNovel_2021}
\bibfield{author}{\bibinfo{person}{Konstantinos Chatzilygeroudis}, \bibinfo{person}{Antoine Cully}, \bibinfo{person}{Vassilis Vassiliades}, {and} \bibinfo{person}{Jean-Baptiste Mouret}.} \bibinfo{year}{2021}\natexlab{}.
\newblock \showarticletitle{Quality-{Diversity} {Optimization}: {A} {Novel} {Branch} of {Stochastic} {Optimization}}.
\newblock In \bibinfo{booktitle}{\emph{Black {Box} {Optimization}, {Machine} {Learning}, and {No}-{Free} {Lunch} {Theorems}}}, \bibfield{editor}{\bibinfo{person}{Panos~M. Pardalos}, \bibinfo{person}{Varvara Rasskazova}, {and} \bibinfo{person}{Michael~N. Vrahatis}} (Eds.). \bibinfo{publisher}{Springer International Publishing}, \bibinfo{address}{Cham}, \bibinfo{pages}{109--135}.
\newblock
\showISBNx{978-3-030-66515-9}
\urldef\tempurl%
\url{https://doi.org/10.1007/978-3-030-66515-9_4}
\showDOI{\tempurl}


\bibitem[Chatzilygeroudis et~al\mbox{.}(2018)]%
        {chatzilygeroudis_ResetfreeTrialandErrorLearning_2018}
\bibfield{author}{\bibinfo{person}{Konstantinos Chatzilygeroudis}, \bibinfo{person}{Vassilis Vassiliades}, {and} \bibinfo{person}{Jean-Baptiste Mouret}.} \bibinfo{year}{2018}\natexlab{}.
\newblock \showarticletitle{Reset-free {Trial}-and-{Error} {Learning} for {Robot} {Damage} {Recovery}}.
\newblock \bibinfo{journal}{\emph{Robotics and Autonomous Systems}}  \bibinfo{volume}{100} (\bibinfo{date}{Feb.} \bibinfo{year}{2018}), \bibinfo{pages}{236--250}.
\newblock
\showISSN{0921-8890}
\urldef\tempurl%
\url{https://doi.org/10.1016/j.robot.2017.11.010}
\showDOI{\tempurl}


\bibitem[Colas et~al\mbox{.}(2020)]%
        {colas_ScalingMAPElitesDeep_2020}
\bibfield{author}{\bibinfo{person}{Cédric Colas}, \bibinfo{person}{Vashisht Madhavan}, \bibinfo{person}{Joost Huizinga}, {and} \bibinfo{person}{Jeff Clune}.} \bibinfo{year}{2020}\natexlab{}.
\newblock \showarticletitle{Scaling {MAP}-{Elites} to deep neuroevolution}. In \bibinfo{booktitle}{\emph{Proceedings of the 2020 {Genetic} and {Evolutionary} {Computation} {Conference}}} \emph{(\bibinfo{series}{{GECCO} '20})}. \bibinfo{publisher}{Association for Computing Machinery}, \bibinfo{address}{New York, NY, USA}, \bibinfo{pages}{67--75}.
\newblock
\showISBNx{978-1-4503-7128-5}
\urldef\tempurl%
\url{https://doi.org/10.1145/3377930.3390217}
\showDOI{\tempurl}


\bibitem[Cully et~al\mbox{.}(2015)]%
        {cully_RobotsThatCan_2015}
\bibfield{author}{\bibinfo{person}{Antoine Cully}, \bibinfo{person}{Jeff Clune}, \bibinfo{person}{Danesh Tarapore}, {and} \bibinfo{person}{Jean-Baptiste Mouret}.} \bibinfo{year}{2015}\natexlab{}.
\newblock \showarticletitle{Robots that can adapt like animals}.
\newblock \bibinfo{journal}{\emph{Nature}} \bibinfo{volume}{521}, \bibinfo{number}{7553} (\bibinfo{date}{May} \bibinfo{year}{2015}), \bibinfo{pages}{503--507}.
\newblock
\showISSN{1476-4687}
\urldef\tempurl%
\url{https://doi.org/10.1038/nature14422}
\showDOI{\tempurl}
\newblock
\shownote{Number: 7553 Publisher: Nature Publishing Group}.


\bibitem[Cully and Demiris(2018)]%
        {cully_QualityDiversityOptimization_2017}
\bibfield{author}{\bibinfo{person}{Antoine Cully} {and} \bibinfo{person}{Yiannis Demiris}.} \bibinfo{year}{2018}\natexlab{}.
\newblock \showarticletitle{Quality and Diversity Optimization: A Unifying Modular Framework}.
\newblock \bibinfo{journal}{\emph{IEEE Transactions on Evolutionary Computation}} \bibinfo{volume}{22}, \bibinfo{number}{2} (\bibinfo{year}{2018}), \bibinfo{pages}{245--259}.
\newblock
\urldef\tempurl%
\url{https://doi.org/10.1109/TEVC.2017.2704781}
\showDOI{\tempurl}


\bibitem[Ecoffet et~al\mbox{.}(2021)]%
        {ecoffet_FirstReturnThen_2021}
\bibfield{author}{\bibinfo{person}{Adrien Ecoffet}, \bibinfo{person}{Joost Huizinga}, \bibinfo{person}{Joel Lehman}, \bibinfo{person}{Kenneth~O. Stanley}, {and} \bibinfo{person}{Jeff Clune}.} \bibinfo{year}{2021}\natexlab{}.
\newblock \showarticletitle{First return, then explore}.
\newblock \bibinfo{journal}{\emph{Nature}} \bibinfo{volume}{590}, \bibinfo{number}{7847} (\bibinfo{date}{Feb.} \bibinfo{year}{2021}), \bibinfo{pages}{580--586}.
\newblock
\showISSN{1476-4687}
\urldef\tempurl%
\url{https://doi.org/10.1038/s41586-020-03157-9}
\showDOI{\tempurl}
\newblock
\shownote{Number: 7847 Publisher: Nature Publishing Group}.


\bibitem[Eysenbach et~al\mbox{.}(2018)]%
        {eysenbach_DiversityAllYou_2018}
\bibfield{author}{\bibinfo{person}{Benjamin Eysenbach}, \bibinfo{person}{Abhishek Gupta}, \bibinfo{person}{Julian Ibarz}, {and} \bibinfo{person}{Sergey Levine}.} \bibinfo{year}{2018}\natexlab{}.
\newblock \bibinfo{title}{Diversity is {All} {You} {Need}: {Learning} {Skills} without a {Reward} {Function}}.
\newblock
\newblock
\urldef\tempurl%
\url{https://doi.org/10.48550/arXiv.1802.06070}
\showDOI{\tempurl}
\newblock
\shownote{arXiv:1802.06070 [cs]}.


\bibitem[Faldor et~al\mbox{.}(2023)]%
        {faldor_dcgme_2023}
\bibfield{author}{\bibinfo{person}{Maxence Faldor}, \bibinfo{person}{F\'{e}lix Chalumeau}, \bibinfo{person}{Manon Flageat}, {and} \bibinfo{person}{Antoine Cully}.} \bibinfo{year}{2023}\natexlab{}.
\newblock \showarticletitle{MAP-Elites with Descriptor-Conditioned Gradients and Archive Distillation into a Single Policy}. In \bibinfo{booktitle}{\emph{Proceedings of the Genetic and Evolutionary Computation Conference}} (Lisbon, Portugal) \emph{(\bibinfo{series}{GECCO '23})}. \bibinfo{publisher}{Association for Computing Machinery}, \bibinfo{address}{New York, NY, USA}, \bibinfo{pages}{138–146}.
\newblock
\showISBNx{9798400701191}
\urldef\tempurl%
\url{https://doi.org/10.1145/3583131.3590503}
\showDOI{\tempurl}


\bibitem[Faldor and Cully(2024)]%
        {faldor2024leniabreeder}
\bibfield{author}{\bibinfo{person}{Maxence Faldor} {and} \bibinfo{person}{Antoine Cully}.} \bibinfo{year}{2024}\natexlab{}.
\newblock \showarticletitle{Toward Artificial Open-Ended Evolution within Lenia using Quality-Diversity}.
\newblock \bibinfo{journal}{\emph{Artificial Life}} (\bibinfo{year}{2024}).
\newblock


\bibitem[Flageat et~al\mbox{.}(2023)]%
        {flageat_EmpiricalAnalysisPGAMAPElites_2023}
\bibfield{author}{\bibinfo{person}{Manon Flageat}, \bibinfo{person}{Félix Chalumeau}, {and} \bibinfo{person}{Antoine Cully}.} \bibinfo{year}{2023}\natexlab{}.
\newblock \showarticletitle{Empirical analysis of {PGA}-{MAP}-{Elites} for {Neuroevolution} in {Uncertain} {Domains}}.
\newblock \bibinfo{journal}{\emph{ACM Transactions on Evolutionary Learning and Optimization}} \bibinfo{volume}{3}, \bibinfo{number}{1} (\bibinfo{date}{March} \bibinfo{year}{2023}), \bibinfo{pages}{1:1--1:32}.
\newblock
\urldef\tempurl%
\url{https://doi.org/10.1145/3577203}
\showDOI{\tempurl}


\bibitem[Flageat and Cully(2023)]%
        {flageat2023uncertain}
\bibfield{author}{\bibinfo{person}{Manon Flageat} {and} \bibinfo{person}{Antoine Cully}.} \bibinfo{year}{2023}\natexlab{}.
\newblock \showarticletitle{Uncertain Quality-Diversity: Evaluation methodology and new methods for Quality-Diversity in Uncertain Domains}.
\newblock \bibinfo{journal}{\emph{IEEE Transactions on Evolutionary Computation}} (\bibinfo{year}{2023}).
\newblock


\bibitem[Flageat et~al\mbox{.}(2022)]%
        {flageat_BenchmarkingQualityDiversityAlgorithms_2022}
\bibfield{author}{\bibinfo{person}{Manon Flageat}, \bibinfo{person}{Bryan Lim}, \bibinfo{person}{Luca Grillotti}, \bibinfo{person}{Maxime Allard}, \bibinfo{person}{Simón~C. Smith}, {and} \bibinfo{person}{Antoine Cully}.} \bibinfo{year}{2022}\natexlab{}.
\newblock \bibinfo{title}{Benchmarking {Quality}-{Diversity} {Algorithms} on {Neuroevolution} for {Reinforcement} {Learning}}.
\newblock
\newblock
\urldef\tempurl%
\url{https://doi.org/10.48550/arXiv.2211.02193}
\showDOI{\tempurl}
\newblock
\shownote{arXiv:2211.02193 [cs]}.


\bibitem[Fontaine and Nikolaidis(2021)]%
        {fontaine_DifferentiableQualityDiversity_2021}
\bibfield{author}{\bibinfo{person}{Matthew Fontaine} {and} \bibinfo{person}{Stefanos Nikolaidis}.} \bibinfo{year}{2021}\natexlab{}.
\newblock \showarticletitle{Differentiable {Quality} {Diversity}}. In \bibinfo{booktitle}{\emph{Advances in {Neural} {Information} {Processing} {Systems}}}, Vol.~\bibinfo{volume}{34}. \bibinfo{publisher}{Curran Associates, Inc.}, \bibinfo{pages}{10040--10052}.
\newblock
\urldef\tempurl%
\url{https://proceedings.neurips.cc/paper/2021/hash/532923f11ac97d3e7cb0130315b067dc-Abstract.html}
\showURL{%
\tempurl}


\bibitem[Fontaine and Nikolaidis(2023)]%
        {fontaine_CovarianceMatrixAdaptation_2023}
\bibfield{author}{\bibinfo{person}{Matthew Fontaine} {and} \bibinfo{person}{Stefanos Nikolaidis}.} \bibinfo{year}{2023}\natexlab{}.
\newblock \showarticletitle{Covariance {Matrix} {Adaptation} {MAP}-{Annealing}}. In \bibinfo{booktitle}{\emph{Proceedings of the {Genetic} and {Evolutionary} {Computation} {Conference}}} \emph{(\bibinfo{series}{{GECCO} '23})}. \bibinfo{publisher}{Association for Computing Machinery}, \bibinfo{address}{New York, NY, USA}, \bibinfo{pages}{456--465}.
\newblock
\showISBNx{9798400701191}
\urldef\tempurl%
\url{https://doi.org/10.1145/3583131.3590389}
\showDOI{\tempurl}


\bibitem[Freeman et~al\mbox{.}(2021)]%
        {brax}
\bibfield{author}{\bibinfo{person}{C.~Daniel Freeman}, \bibinfo{person}{Erik Frey}, \bibinfo{person}{Anton Raichuk}, \bibinfo{person}{Sertan Girgin}, \bibinfo{person}{Igor Mordatch}, {and} \bibinfo{person}{Olivier Bachem}.} \bibinfo{year}{2021}\natexlab{}.
\newblock \bibinfo{booktitle}{\emph{Brax - A Differentiable Physics Engine for Large Scale Rigid Body Simulation}}.
\newblock
\urldef\tempurl%
\url{http://github.com/google/brax}
\showURL{%
\tempurl}


\bibitem[Fujimoto et~al\mbox{.}(2018)]%
        {fujimoto_AddressingFunctionApproximation_2018}
\bibfield{author}{\bibinfo{person}{Scott Fujimoto}, \bibinfo{person}{Herke Hoof}, {and} \bibinfo{person}{David Meger}.} \bibinfo{year}{2018}\natexlab{}.
\newblock \showarticletitle{Addressing {Function} {Approximation} {Error} in {Actor}-{Critic} {Methods}}. In \bibinfo{booktitle}{\emph{Proceedings of the 35th {International} {Conference} on {Machine} {Learning}}}. \bibinfo{publisher}{PMLR}, \bibinfo{pages}{1587--1596}.
\newblock
\urldef\tempurl%
\url{https://proceedings.mlr.press/v80/fujimoto18a.html}
\showURL{%
\tempurl}
\newblock
\shownote{ISSN: 2640-3498}.


\bibitem[Gregor et~al\mbox{.}(2016)]%
        {gregor_VariationalIntrinsicControl_2016}
\bibfield{author}{\bibinfo{person}{Karol Gregor}, \bibinfo{person}{Danilo~Jimenez Rezende}, {and} \bibinfo{person}{Daan Wierstra}.} \bibinfo{year}{2016}\natexlab{}.
\newblock \bibinfo{title}{Variational {Intrinsic} {Control}}.
\newblock
\newblock
\urldef\tempurl%
\url{https://doi.org/10.48550/arXiv.1611.07507}
\showDOI{\tempurl}
\newblock
\shownote{arXiv:1611.07507 [cs]}.


\bibitem[Grillotti et~al\mbox{.}(2024)]%
        {airl2024qdac}
\bibfield{author}{\bibinfo{person}{Luca Grillotti}, \bibinfo{person}{Maxence Faldor}, \bibinfo{person}{Borja González~León}, {and} \bibinfo{person}{Antoine Cully}.} \bibinfo{year}{2024}\natexlab{}.
\newblock \showarticletitle{Quality-Diversity Actor-Critic: Learning High-Performing and Diverse Behaviors via Value and Successor Features Critics}. In \bibinfo{booktitle}{\emph{International Conference on Machine Learning}}. PMLR.
\newblock


\bibitem[Haarnoja et~al\mbox{.}(2018)]%
        {haarnoja_SoftActorCriticOffPolicy_2018}
\bibfield{author}{\bibinfo{person}{Tuomas Haarnoja}, \bibinfo{person}{Aurick Zhou}, \bibinfo{person}{Pieter Abbeel}, {and} \bibinfo{person}{Sergey Levine}.} \bibinfo{year}{2018}\natexlab{}.
\newblock \showarticletitle{Soft {Actor}-{Critic}: {Off}-{Policy} {Maximum} {Entropy} {Deep} {Reinforcement} {Learning} with a {Stochastic} {Actor}}. In \bibinfo{booktitle}{\emph{Proceedings of the 35th {International} {Conference} on {Machine} {Learning}}}. \bibinfo{publisher}{PMLR}, \bibinfo{pages}{1861--1870}.
\newblock
\urldef\tempurl%
\url{https://proceedings.mlr.press/v80/haarnoja18b.html}
\showURL{%
\tempurl}
\newblock
\shownote{ISSN: 2640-3498}.


\bibitem[Haarnoja et~al\mbox{.}(2019)]%
        {haarnoja_SoftActorCriticAlgorithms_2019}
\bibfield{author}{\bibinfo{person}{Tuomas Haarnoja}, \bibinfo{person}{Aurick Zhou}, \bibinfo{person}{Kristian Hartikainen}, \bibinfo{person}{George Tucker}, \bibinfo{person}{Sehoon Ha}, \bibinfo{person}{Jie Tan}, \bibinfo{person}{Vikash Kumar}, \bibinfo{person}{Henry Zhu}, \bibinfo{person}{Abhishek Gupta}, \bibinfo{person}{Pieter Abbeel}, {and} \bibinfo{person}{Sergey Levine}.} \bibinfo{year}{2019}\natexlab{}.
\newblock \bibinfo{title}{Soft {Actor}-{Critic} {Algorithms} and {Applications}}.
\newblock
\newblock
\urldef\tempurl%
\url{https://doi.org/10.48550/arXiv.1812.05905}
\showDOI{\tempurl}
\newblock
\shownote{arXiv:1812.05905 [cs, stat]}.


\bibitem[Hansen(2023)]%
        {hansen_CMAEvolutionStrategy_2023}
\bibfield{author}{\bibinfo{person}{Nikolaus Hansen}.} \bibinfo{year}{2023}\natexlab{}.
\newblock \bibinfo{title}{The {CMA} {Evolution} {Strategy}: {A} {Tutorial}}.
\newblock
\newblock
\urldef\tempurl%
\url{https://doi.org/10.48550/arXiv.1604.00772}
\showDOI{\tempurl}
\newblock
\shownote{arXiv:1604.00772 [cs, stat]}.


\bibitem[Heess et~al\mbox{.}(2017)]%
        {heess_EmergenceLocomotionBehaviours_2017}
\bibfield{author}{\bibinfo{person}{Nicolas Heess}, \bibinfo{person}{Dhruva TB}, \bibinfo{person}{Srinivasan Sriram}, \bibinfo{person}{Jay Lemmon}, \bibinfo{person}{Josh Merel}, \bibinfo{person}{Greg Wayne}, \bibinfo{person}{Yuval Tassa}, \bibinfo{person}{Tom Erez}, \bibinfo{person}{Ziyu Wang}, \bibinfo{person}{Ali Eslami}, \bibinfo{person}{Martin Riedmiller}, {and} \bibinfo{person}{David Silver}.} \bibinfo{year}{2017}\natexlab{}.
\newblock \showarticletitle{Emergence of {Locomotion} {Behaviours} in {Rich} {Environments}}.
\newblock  (\bibinfo{date}{July} \bibinfo{year}{2017}).
\newblock


\bibitem[Hornik et~al\mbox{.}(1989)]%
        {hornik_MultilayerFeedforwardNetworks_1989}
\bibfield{author}{\bibinfo{person}{Kurt Hornik}, \bibinfo{person}{Maxwell Stinchcombe}, {and} \bibinfo{person}{Halbert White}.} \bibinfo{year}{1989}\natexlab{}.
\newblock \showarticletitle{Multilayer feedforward networks are universal approximators}.
\newblock \bibinfo{journal}{\emph{Neural Networks}} \bibinfo{volume}{2}, \bibinfo{number}{5} (\bibinfo{year}{1989}), \bibinfo{pages}{359--366}.
\newblock
\showISSN{0893-6080}
\urldef\tempurl%
\url{https://doi.org/10.1016/0893-6080(89)90020-8}
\showDOI{\tempurl}


\bibitem[Jaderberg et~al\mbox{.}(2017)]%
        {jaderberg_PopulationBasedTraining_2017}
\bibfield{author}{\bibinfo{person}{Max Jaderberg}, \bibinfo{person}{Valentin Dalibard}, \bibinfo{person}{Simon Osindero}, \bibinfo{person}{Wojciech~M. Czarnecki}, \bibinfo{person}{Jeff Donahue}, \bibinfo{person}{Ali Razavi}, \bibinfo{person}{Oriol Vinyals}, \bibinfo{person}{Tim Green}, \bibinfo{person}{Iain Dunning}, \bibinfo{person}{Karen Simonyan}, \bibinfo{person}{Chrisantha Fernando}, {and} \bibinfo{person}{Koray Kavukcuoglu}.} \bibinfo{year}{2017}\natexlab{}.
\newblock \bibinfo{title}{Population {Based} {Training} of {Neural} {Networks}}.
\newblock
\newblock
\urldef\tempurl%
\url{https://doi.org/10.48550/arXiv.1711.09846}
\showDOI{\tempurl}
\newblock
\shownote{arXiv:1711.09846 [cs]}.


\bibitem[Jegorova et~al\mbox{.}(2019)]%
        {jegorova_BehaviouralRepertoireGenerative_2019}
\bibfield{author}{\bibinfo{person}{Marija Jegorova}, \bibinfo{person}{Stéphane Doncieux}, {and} \bibinfo{person}{Timothy Hospedales}.} \bibinfo{year}{2019}\natexlab{}.
\newblock \showarticletitle{Behavioural {Repertoire} via {Generative} {Adversarial} {Policy} {Networks}}. In \bibinfo{booktitle}{\emph{2019 {Joint} {IEEE} 9th {International} {Conference} on {Development} and {Learning} and {Epigenetic} {Robotics} ({ICDL}-{EpiRob})}}.
\newblock
\urldef\tempurl%
\url{https://doi.org/10.1109/ICDL-EpiRob44920.2019}
\showDOI{\tempurl}
\newblock
\shownote{arXiv:1811.02945 [cs, stat]}.


\bibitem[Kumar et~al\mbox{.}(2020)]%
        {kumar_OneSolutionNot_2020}
\bibfield{author}{\bibinfo{person}{Saurabh Kumar}, \bibinfo{person}{Aviral Kumar}, \bibinfo{person}{Sergey Levine}, {and} \bibinfo{person}{Chelsea Finn}.} \bibinfo{year}{2020}\natexlab{}.
\newblock \showarticletitle{One {Solution} is {Not} {All} {You} {Need}: {Few}-{Shot} {Extrapolation} via {Structured} {MaxEnt} {RL}}. In \bibinfo{booktitle}{\emph{Advances in {Neural} {Information} {Processing} {Systems}}}, Vol.~\bibinfo{volume}{33}. \bibinfo{publisher}{Curran Associates, Inc.}, \bibinfo{pages}{8198--8210}.
\newblock
\urldef\tempurl%
\url{https://proceedings.neurips.cc/paper/2020/hash/5d151d1059a6281335a10732fc49620e-Abstract.html}
\showURL{%
\tempurl}


\bibitem[Lillicrap et~al\mbox{.}(2016)]%
        {lillicrap_ContinuousControlDeep_2019}
\bibfield{author}{\bibinfo{person}{Timothy~P. Lillicrap}, \bibinfo{person}{Jonathan~J. Hunt}, \bibinfo{person}{Alexander Pritzel}, \bibinfo{person}{Nicolas Heess}, \bibinfo{person}{Tom Erez}, \bibinfo{person}{Yuval Tassa}, \bibinfo{person}{David Silver}, {and} \bibinfo{person}{Daan Wierstra}.} \bibinfo{year}{2016}\natexlab{}.
\newblock \showarticletitle{Continuous control with deep reinforcement learning}. In \bibinfo{booktitle}{\emph{4th International Conference on Learning Representations, {ICLR} 2016, San Juan, Puerto Rico, May 2-4, 2016, Conference Track Proceedings}}, \bibfield{editor}{\bibinfo{person}{Yoshua Bengio} {and} \bibinfo{person}{Yann LeCun}} (Eds.).
\newblock
\urldef\tempurl%
\url{http://arxiv.org/abs/1509.02971}
\showURL{%
\tempurl}


\bibitem[Mnih et~al\mbox{.}(2013)]%
        {mnih_PlayingAtariDeep_2013}
\bibfield{author}{\bibinfo{person}{Volodymyr Mnih}, \bibinfo{person}{Koray Kavukcuoglu}, \bibinfo{person}{David Silver}, \bibinfo{person}{Alex Graves}, \bibinfo{person}{Ioannis Antonoglou}, \bibinfo{person}{Daan Wierstra}, {and} \bibinfo{person}{Martin Riedmiller}.} \bibinfo{year}{2013}\natexlab{}.
\newblock \bibinfo{title}{Playing {Atari} with {Deep} {Reinforcement} {Learning}}.
\newblock
\newblock
\urldef\tempurl%
\url{https://doi.org/10.48550/arXiv.1312.5602}
\showDOI{\tempurl}
\newblock
\shownote{arXiv:1312.5602 [cs]}.


\bibitem[Mnih et~al\mbox{.}(2015)]%
        {mnih_HumanlevelControlDeep_2015}
\bibfield{author}{\bibinfo{person}{Volodymyr Mnih}, \bibinfo{person}{Koray Kavukcuoglu}, \bibinfo{person}{David Silver}, \bibinfo{person}{Andrei~A. Rusu}, \bibinfo{person}{Joel Veness}, \bibinfo{person}{Marc~G. Bellemare}, \bibinfo{person}{Alex Graves}, \bibinfo{person}{Martin Riedmiller}, \bibinfo{person}{Andreas~K. Fidjeland}, \bibinfo{person}{Georg Ostrovski}, \bibinfo{person}{Stig Petersen}, \bibinfo{person}{Charles Beattie}, \bibinfo{person}{Amir Sadik}, \bibinfo{person}{Ioannis Antonoglou}, \bibinfo{person}{Helen King}, \bibinfo{person}{Dharshan Kumaran}, \bibinfo{person}{Daan Wierstra}, \bibinfo{person}{Shane Legg}, {and} \bibinfo{person}{Demis Hassabis}.} \bibinfo{year}{2015}\natexlab{}.
\newblock \showarticletitle{Human-level control through deep reinforcement learning}.
\newblock \bibinfo{journal}{\emph{Nature}} \bibinfo{volume}{518}, \bibinfo{number}{7540} (\bibinfo{date}{Feb.} \bibinfo{year}{2015}), \bibinfo{pages}{529--533}.
\newblock
\showISSN{1476-4687}
\urldef\tempurl%
\url{https://doi.org/10.1038/nature14236}
\showDOI{\tempurl}
\newblock
\shownote{Number: 7540 Publisher: Nature Publishing Group}.


\bibitem[Mouret and Clune(2015)]%
        {mouret_IlluminatingSearchSpaces_2015}
\bibfield{author}{\bibinfo{person}{Jean{-}Baptiste Mouret} {and} \bibinfo{person}{Jeff Clune}.} \bibinfo{year}{2015}\natexlab{}.
\newblock \showarticletitle{Illuminating search spaces by mapping elites}.
\newblock \bibinfo{journal}{\emph{CoRR}}  \bibinfo{volume}{abs/1504.04909} (\bibinfo{year}{2015}).
\newblock
\showeprint[arXiv]{1504.04909}
\urldef\tempurl%
\url{http://arxiv.org/abs/1504.04909}
\showURL{%
\tempurl}


\bibitem[Nilsson and Cully(2021)]%
        {nilsson_PolicyGradientAssisted_2021}
\bibfield{author}{\bibinfo{person}{Olle Nilsson} {and} \bibinfo{person}{Antoine Cully}.} \bibinfo{year}{2021}\natexlab{}.
\newblock \showarticletitle{Policy gradient assisted {MAP}-{Elites}}. In \bibinfo{booktitle}{\emph{Proceedings of the {Genetic} and {Evolutionary} {Computation} {Conference}}} \emph{(\bibinfo{series}{{GECCO} '21})}. \bibinfo{publisher}{Association for Computing Machinery}, \bibinfo{address}{New York, NY, USA}, \bibinfo{pages}{866--875}.
\newblock
\showISBNx{978-1-4503-8350-9}
\urldef\tempurl%
\url{https://doi.org/10.1145/3449639.3459304}
\showDOI{\tempurl}


\bibitem[OpenAI et~al\mbox{.}(2019)]%
        {openai_SolvingRubikCube_2019}
\bibfield{author}{\bibinfo{person}{OpenAI}, \bibinfo{person}{Ilge Akkaya}, \bibinfo{person}{Marcin Andrychowicz}, \bibinfo{person}{Maciek Chociej}, \bibinfo{person}{Mateusz Litwin}, \bibinfo{person}{Bob McGrew}, \bibinfo{person}{Arthur Petron}, \bibinfo{person}{Alex Paino}, \bibinfo{person}{Matthias Plappert}, \bibinfo{person}{Glenn Powell}, \bibinfo{person}{Raphael Ribas}, \bibinfo{person}{Jonas Schneider}, \bibinfo{person}{Nikolas Tezak}, \bibinfo{person}{Jerry Tworek}, \bibinfo{person}{Peter Welinder}, \bibinfo{person}{Lilian Weng}, \bibinfo{person}{Qiming Yuan}, \bibinfo{person}{Wojciech Zaremba}, {and} \bibinfo{person}{Lei Zhang}.} \bibinfo{year}{2019}\natexlab{}.
\newblock \bibinfo{title}{Solving {Rubik}'s {Cube} with a {Robot} {Hand}}.
\newblock
\newblock
\urldef\tempurl%
\url{https://doi.org/10.48550/arXiv.1910.07113}
\showDOI{\tempurl}
\newblock
\shownote{arXiv:1910.07113 [cs, stat]}.


\bibitem[Ostrovski et~al\mbox{.}(2021)]%
        {ostrovski_difficulty_2021}
\bibfield{author}{\bibinfo{person}{Georg Ostrovski}, \bibinfo{person}{Pablo~Samuel Castro}, {and} \bibinfo{person}{Will Dabney}.} \bibinfo{year}{2021}\natexlab{}.
\newblock \bibinfo{title}{The {Difficulty} of {Passive} {Learning} in {Deep} {Reinforcement} {Learning}}.
\newblock
\newblock
\urldef\tempurl%
\url{http://arxiv.org/abs/2110.14020}
\showURL{%
\tempurl}
\newblock
\shownote{arXiv:2110.14020 [cs]}.


\bibitem[Pierrot and Flajolet(2023)]%
        {pierrot_EvolvingPopulationsDiverse_2023}
\bibfield{author}{\bibinfo{person}{Thomas Pierrot} {and} \bibinfo{person}{Arthur Flajolet}.} \bibinfo{year}{2023}\natexlab{}.
\newblock \bibinfo{title}{Evolving {Populations} of {Diverse} {RL} {Agents} with {MAP}-{Elites}}.
\newblock
\newblock
\urldef\tempurl%
\url{https://doi.org/10.48550/arXiv.2303.12803}
\showDOI{\tempurl}
\newblock
\shownote{arXiv:2303.12803 [cs]}.


\bibitem[Pierrot et~al\mbox{.}(2022)]%
        {pierrot_DiversityPolicyGradient_2022}
\bibfield{author}{\bibinfo{person}{Thomas Pierrot}, \bibinfo{person}{Valentin Macé}, \bibinfo{person}{Felix Chalumeau}, \bibinfo{person}{Arthur Flajolet}, \bibinfo{person}{Geoffrey Cideron}, \bibinfo{person}{Karim Beguir}, \bibinfo{person}{Antoine Cully}, \bibinfo{person}{Olivier Sigaud}, {and} \bibinfo{person}{Nicolas Perrin-Gilbert}.} \bibinfo{year}{2022}\natexlab{}.
\newblock \showarticletitle{Diversity policy gradient for sample efficient quality-diversity optimization}. In \bibinfo{booktitle}{\emph{Proceedings of the {Genetic} and {Evolutionary} {Computation} {Conference}}} \emph{(\bibinfo{series}{{GECCO} '22})}. \bibinfo{publisher}{Association for Computing Machinery}, \bibinfo{address}{New York, NY, USA}, \bibinfo{pages}{1075--1083}.
\newblock
\showISBNx{978-1-4503-9237-2}
\urldef\tempurl%
\url{https://doi.org/10.1145/3512290.3528845}
\showDOI{\tempurl}


\bibitem[Pugh et~al\mbox{.}(2016)]%
        {pugh_QualityDiversityNew_2016}
\bibfield{author}{\bibinfo{person}{Justin~K. Pugh}, \bibinfo{person}{Lisa~B. Soros}, {and} \bibinfo{person}{Kenneth~O. Stanley}.} \bibinfo{year}{2016}\natexlab{}.
\newblock \showarticletitle{Quality {Diversity}: {A} {New} {Frontier} for {Evolutionary} {Computation}}.
\newblock \bibinfo{journal}{\emph{Frontiers in Robotics and AI}}  \bibinfo{volume}{3} (\bibinfo{year}{2016}).
\newblock
\showISSN{2296-9144}
\urldef\tempurl%
\url{https://www.frontiersin.org/articles/10.3389/frobt.2016.00040}
\showURL{%
\tempurl}


\bibitem[Salimans et~al\mbox{.}(2017)]%
        {salimans2017evolution}
\bibfield{author}{\bibinfo{person}{Tim Salimans}, \bibinfo{person}{Jonathan Ho}, \bibinfo{person}{Xi Chen}, \bibinfo{person}{Szymon Sidor}, {and} \bibinfo{person}{Ilya Sutskever}.} \bibinfo{year}{2017}\natexlab{}.
\newblock \showarticletitle{Evolution strategies as a scalable alternative to reinforcement learning}.
\newblock \bibinfo{journal}{\emph{arXiv preprint arXiv:1703.03864}} (\bibinfo{year}{2017}).
\newblock


\bibitem[Schaul et~al\mbox{.}(2015)]%
        {schaul_UniversalValueFunction_2015}
\bibfield{author}{\bibinfo{person}{Tom Schaul}, \bibinfo{person}{Daniel Horgan}, \bibinfo{person}{Karol Gregor}, {and} \bibinfo{person}{David Silver}.} \bibinfo{year}{2015}\natexlab{}.
\newblock \showarticletitle{Universal {Value} {Function} {Approximators}}. In \bibinfo{booktitle}{\emph{Proceedings of the 32nd {International} {Conference} on {Machine} {Learning}}}. \bibinfo{publisher}{PMLR}, \bibinfo{pages}{1312--1320}.
\newblock
\urldef\tempurl%
\url{https://proceedings.mlr.press/v37/schaul15.html}
\showURL{%
\tempurl}
\newblock
\shownote{ISSN: 1938-7228}.


\bibitem[Sharma et~al\mbox{.}(2019)]%
        {sharma_DynamicsAwareUnsupervisedDiscovery_2019}
\bibfield{author}{\bibinfo{person}{Archit Sharma}, \bibinfo{person}{Shixiang Gu}, \bibinfo{person}{Sergey Levine}, \bibinfo{person}{Vikash Kumar}, {and} \bibinfo{person}{Karol Hausman}.} \bibinfo{year}{2019}\natexlab{}.
\newblock \showarticletitle{Dynamics-{Aware} {Unsupervised} {Discovery} of {Skills}}.
\newblock
\urldef\tempurl%
\url{https://openreview.net/forum?id=HJgLZR4KvH}
\showURL{%
\tempurl}


\bibitem[Silver et~al\mbox{.}(2016)]%
        {silver_MasteringGameGo_2016}
\bibfield{author}{\bibinfo{person}{David Silver}, \bibinfo{person}{Aja Huang}, \bibinfo{person}{Chris~J. Maddison}, \bibinfo{person}{Arthur Guez}, \bibinfo{person}{Laurent Sifre}, \bibinfo{person}{George van~den Driessche}, \bibinfo{person}{Julian Schrittwieser}, \bibinfo{person}{Ioannis Antonoglou}, \bibinfo{person}{Veda Panneershelvam}, \bibinfo{person}{Marc Lanctot}, \bibinfo{person}{Sander Dieleman}, \bibinfo{person}{Dominik Grewe}, \bibinfo{person}{John Nham}, \bibinfo{person}{Nal Kalchbrenner}, \bibinfo{person}{Ilya Sutskever}, \bibinfo{person}{Timothy Lillicrap}, \bibinfo{person}{Madeleine Leach}, \bibinfo{person}{Koray Kavukcuoglu}, \bibinfo{person}{Thore Graepel}, {and} \bibinfo{person}{Demis Hassabis}.} \bibinfo{year}{2016}\natexlab{}.
\newblock \showarticletitle{Mastering the game of {Go} with deep neural networks and tree search}.
\newblock \bibinfo{journal}{\emph{Nature}} \bibinfo{volume}{529}, \bibinfo{number}{7587} (\bibinfo{date}{Jan.} \bibinfo{year}{2016}), \bibinfo{pages}{484--489}.
\newblock
\showISSN{1476-4687}
\urldef\tempurl%
\url{https://doi.org/10.1038/nature16961}
\showDOI{\tempurl}
\newblock
\shownote{Number: 7587 Publisher: Nature Publishing Group}.


\bibitem[Silver et~al\mbox{.}(2014)]%
        {silver_DeterministicPolicyGradient_2014}
\bibfield{author}{\bibinfo{person}{David Silver}, \bibinfo{person}{Guy Lever}, \bibinfo{person}{Nicolas Heess}, \bibinfo{person}{Thomas Degris}, \bibinfo{person}{Daan Wierstra}, {and} \bibinfo{person}{Martin Riedmiller}.} \bibinfo{year}{2014}\natexlab{}.
\newblock \showarticletitle{Deterministic {Policy} {Gradient} {Algorithms}}. In \bibinfo{booktitle}{\emph{Proceedings of the 31st {International} {Conference} on {Machine} {Learning}}}. \bibinfo{publisher}{PMLR}, \bibinfo{pages}{387--395}.
\newblock
\urldef\tempurl%
\url{https://proceedings.mlr.press/v32/silver14.html}
\showURL{%
\tempurl}
\newblock
\shownote{ISSN: 1938-7228}.


\bibitem[Sutton and Barto(2018)]%
        {sutton_ReinforcementLearningIntroduction_2018}
\bibfield{author}{\bibinfo{person}{Richard~S. Sutton} {and} \bibinfo{person}{Andrew~G. Barto}.} \bibinfo{year}{2018}\natexlab{}.
\newblock \bibinfo{booktitle}{\emph{Reinforcement learning: an introduction} (\bibinfo{edition}{second edition} ed.)}.
\newblock \bibinfo{publisher}{The MIT Press}, \bibinfo{address}{Cambridge, Massachusetts}.
\newblock
\showISBNx{978-0-262-03924-6}


\bibitem[Tjanaka et~al\mbox{.}(2023)]%
        {tjanaka_TrainingDiverseHighDimensional_2023}
\bibfield{author}{\bibinfo{person}{Bryon Tjanaka}, \bibinfo{person}{Matthew~C. Fontaine}, \bibinfo{person}{David~H. Lee}, \bibinfo{person}{Aniruddha Kalkar}, {and} \bibinfo{person}{Stefanos Nikolaidis}.} \bibinfo{year}{2023}\natexlab{}.
\newblock \bibinfo{title}{Training {Diverse} {High}-{Dimensional} {Controllers} by {Scaling} {Covariance} {Matrix} {Adaptation} {MAP}-{Annealing}}.
\newblock
\newblock
\urldef\tempurl%
\url{https://doi.org/10.48550/arXiv.2210.02622}
\showDOI{\tempurl}
\newblock
\shownote{arXiv:2210.02622 [cs]}.


\bibitem[Tjanaka et~al\mbox{.}(2022)]%
        {tjanaka_ApproximatingGradientsDifferentiable_2022}
\bibfield{author}{\bibinfo{person}{Bryon Tjanaka}, \bibinfo{person}{Matthew~C. Fontaine}, \bibinfo{person}{Julian Togelius}, {and} \bibinfo{person}{Stefanos Nikolaidis}.} \bibinfo{year}{2022}\natexlab{}.
\newblock \showarticletitle{Approximating gradients for differentiable quality diversity in reinforcement learning}. In \bibinfo{booktitle}{\emph{Proceedings of the {Genetic} and {Evolutionary} {Computation} {Conference}}} \emph{(\bibinfo{series}{{GECCO} '22})}. \bibinfo{publisher}{Association for Computing Machinery}, \bibinfo{address}{New York, NY, USA}, \bibinfo{pages}{1102--1111}.
\newblock
\showISBNx{978-1-4503-9237-2}
\urldef\tempurl%
\url{https://doi.org/10.1145/3512290.3528705}
\showDOI{\tempurl}


\bibitem[Vassiliades et~al\mbox{.}(2018)]%
        {vassiliades_UsingCentroidalVoronoi_2017}
\bibfield{author}{\bibinfo{person}{Vassilis Vassiliades}, \bibinfo{person}{Konstantinos Chatzilygeroudis}, {and} \bibinfo{person}{Jean-Baptiste Mouret}.} \bibinfo{year}{2018}\natexlab{}.
\newblock \showarticletitle{Using Centroidal Voronoi Tessellations to Scale Up the Multidimensional Archive of Phenotypic Elites Algorithm}.
\newblock \bibinfo{journal}{\emph{IEEE Transactions on Evolutionary Computation}} \bibinfo{volume}{22}, \bibinfo{number}{4} (\bibinfo{year}{2018}), \bibinfo{pages}{623--630}.
\newblock
\urldef\tempurl%
\url{https://doi.org/10.1109/TEVC.2017.2735550}
\showDOI{\tempurl}


\end{thebibliography}

\clearpage
\appendix

\newcommand{\MyTag}{appendix}
\newpage
\section{Supplementary Results}
\subsection{Archives}
\label{appendix:archives}
We provide the archives obtained at the end of training for each algorithm on all environments. For each (algorithm, environment) pair, we select the most representative seed with the QD score closest to the median QD score over all seeds to avoid cherry picking.

\begin{figure}[H]
\centering
\includegraphics[width=\textwidth]{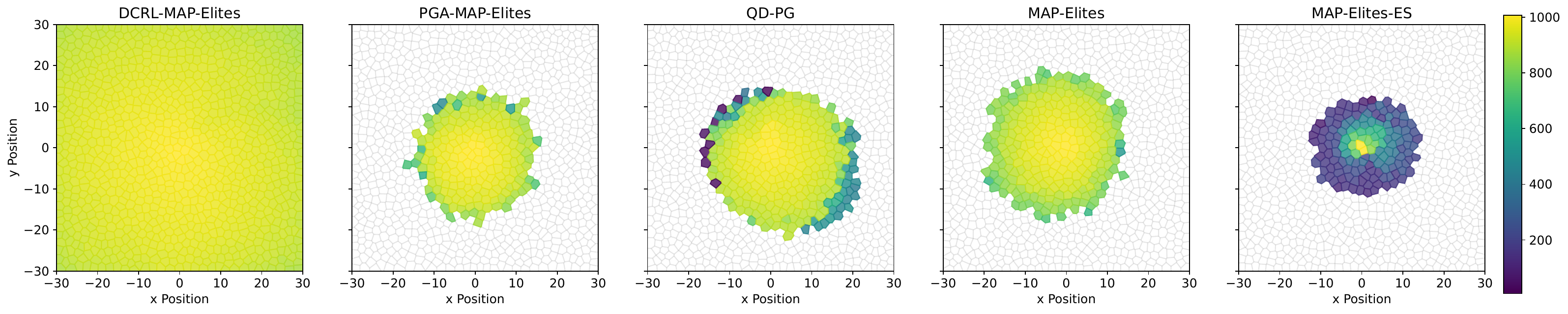}
\caption{\textbf{Ant Omni} Archive at the end of training for all algorithms.}
\Description{\textbf{Ant Omni} Archive at the end of training for all algorithms.}
\end{figure}
\begin{figure}[H]
\centering
\includegraphics[width=\textwidth]{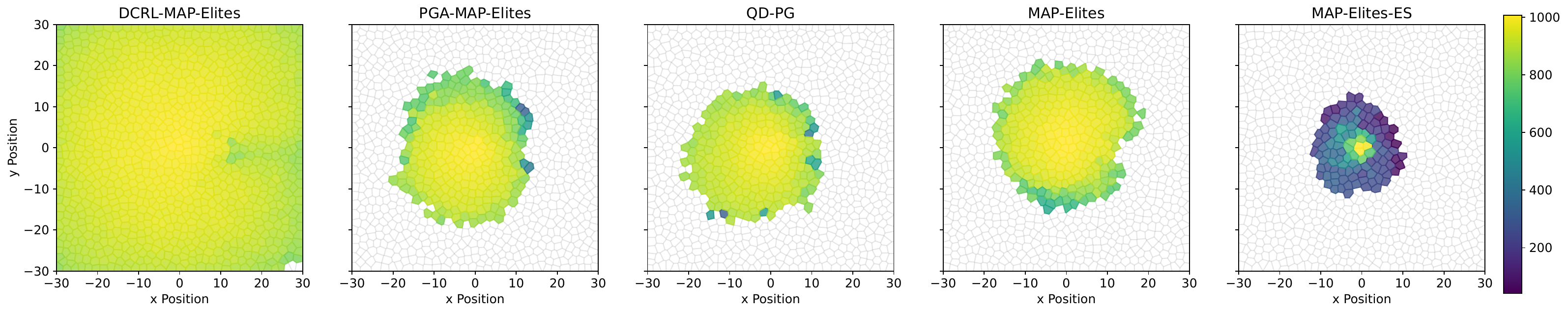}
\caption{\textbf{AntTrap Omni} Archive at the end of training for all algorithms.}
\Description{\textbf{AntTrap Omni} Archive at the end of training for all algorithms.}
\end{figure}
\begin{figure}[H]
\centering
\includegraphics[width=\textwidth]{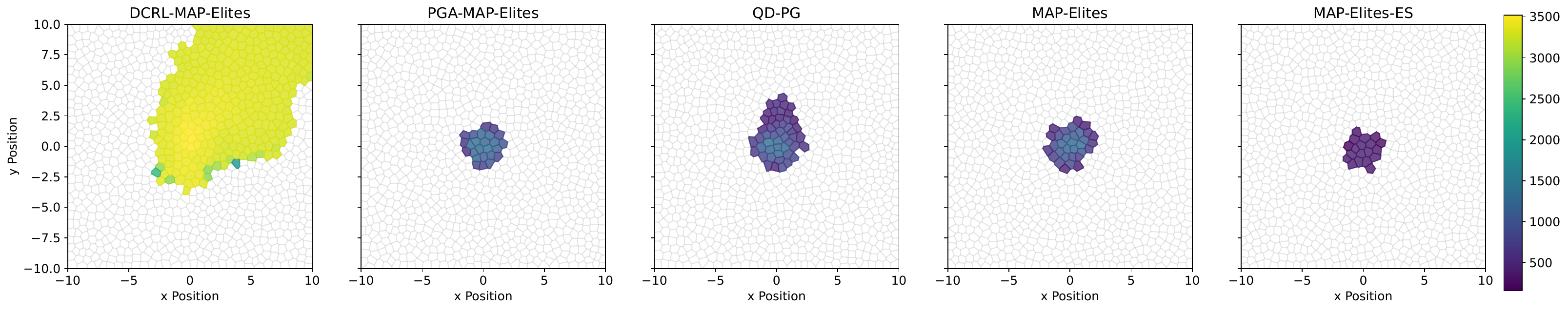}
\caption{\textbf{Humanoid Omni} Archive at the end of training for all algorithms.}
\Description{\textbf{Humanoid Omni} Archive at the end of training for all algorithms.}
\end{figure}
\begin{figure}[H]
\centering
\includegraphics[width=\textwidth]{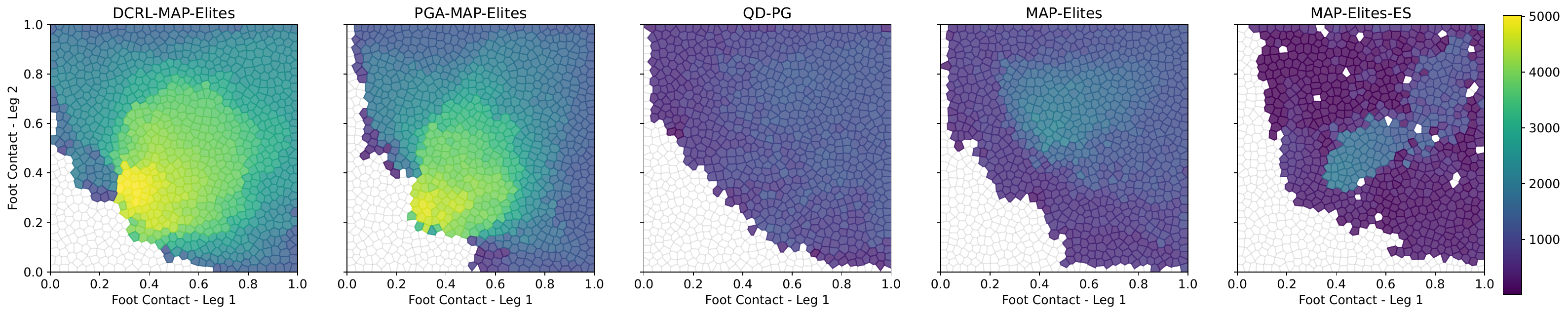}
\caption{\textbf{Walker Uni} Archive at the end of training for all algorithms.}
\Description{\textbf{Walker Uni} Archive at the end of training for all algorithms.}
\end{figure}
\begin{figure}[H]
\centering
\includegraphics[width=\textwidth]{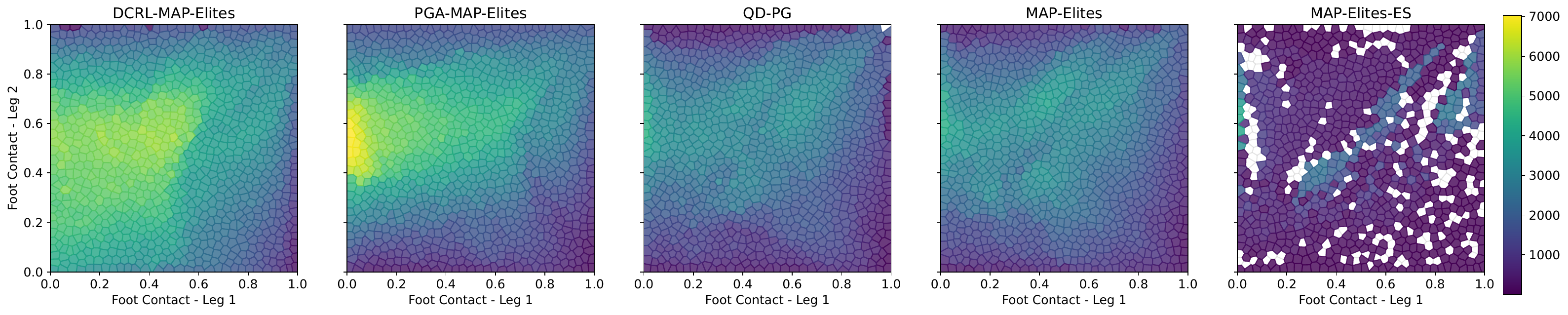}
\caption{\textbf{Halfcheetah Uni} Archive at the end of training for all algorithms.}
\Description{\textbf{Halfcheetah Uni} Archive at the end of training for all algorithms.}
\end{figure}
\begin{figure}[H]
\centering
\includegraphics[width=\textwidth]{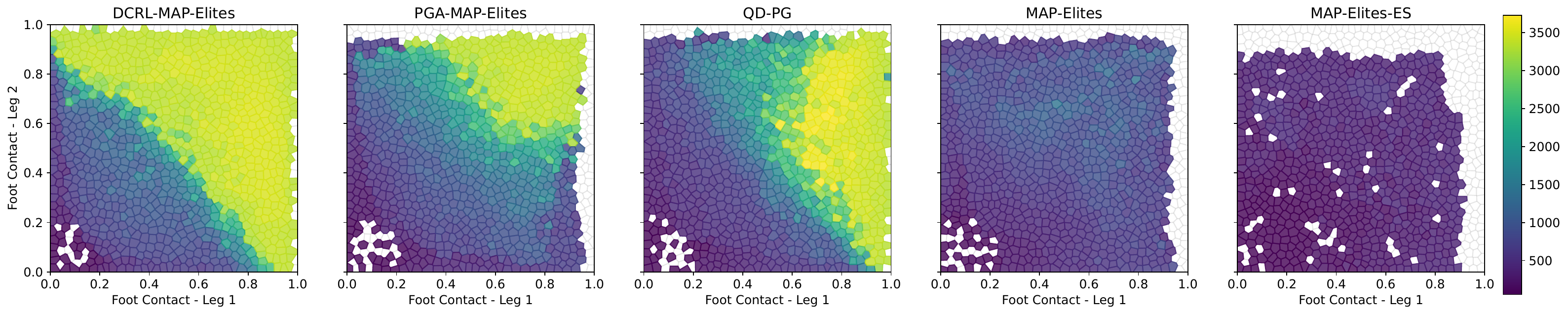}
\caption{\textbf{Humanoid Uni} Archive at the end of training for all algorithms.}
\Description{\textbf{Humanoid Uni} Archive at the end of training for all algorithms.}
\end{figure}

\newpage
\section{Algorithms}
\subsection{\dcrlme{}}

\begin{algorithm}[H]
\caption{Descriptor-Conditioned Actor-Critic Training}
\MyLabel{alg:dcrl-me-actor-critic}
\begin{algorithmic}
\Function{\textsc{train\_actor\_critic}}{$\policy_\actorParams, \critic_{\criticParams_1}, \critic_{\criticParams_2}, \replayBuffer$}
    \For{$t = 1  \rightarrow \ncritic$}
        \State Sample $N$ transitions $\left(\obs, \action, \reward, \obs^\prime, \desc, \desc^\prime \right)$ from $\replayBuffer$
        \State Sample smoothing noise $\epsilon$
        \State $y \gets S(\desc, \desc^\prime) \, \reward + \discount \min\limits_{i=1,2}  \critic_{\criticParams_{i}^\prime}(\obs^\prime, \policy_{\actorParams^\prime}(\obs^\prime \mid \desc^\prime) + \epsilon \mid \desc^\prime)$
        \State Update both critics by regression to $y$
        \If{$t$ mod $\delay$}
            \State Update actor using the deterministic policy gradient:
            \State 	$\frac{1}{N}\sum \nabla_\actorParams \policy_{\actorParams}(\obs \mid \desc^\prime) \nabla_\action \critic_{\criticParams_1}(\obs, \action \mid \desc^\prime)|_{\action=\policy_{\actorParams}(\obs \mid \desc^\prime)}$
            \State Soft-update target networks $\critic_{\criticParams{i}^\prime}$ and $\policy_{\actorParams^\prime}$
        \EndIf
    \EndFor
\EndFunction
\end{algorithmic}
\end{algorithm}

\begin{algorithm}[H]
\caption{Descriptor-Conditioned PG Variation}
\MyLabel{alg:dcrl-me-qpg}
\begin{algorithmic}
\Function{\textsc{variation\_pg}}{$\policy_{\policyParams} \dots,  \critic_{\criticParams_1}, \replayBuffer$}
    \For{$\policy_{\policyParams} \dots$}
        \State $\desc_\policyParams \gets \descFunction(\policy_{\policyParams})$
        \For{$i = 1  \rightarrow \npg$}
            \State Sample $N$ transitions $\left(\obs, \action, \reward, \obs^\prime, \desc, \desc^\prime \right)$ from $\replayBuffer$
            \State Update actor using the deterministic policy gradient:
            \State 	$\frac{1}{N}\sum \nabla_\policyParams \policy_{\policyParams}(\obs) \nabla_\action \critic_{\criticParams_1}(\obs, \action \mid \desc_\policyParams)|_{\action=\policy_{\policyParams}(\obs)}$
        \EndFor
    \EndFor
    \State \Return $\policy_{\widehat{\actorParams}} \dots$
\EndFunction
\end{algorithmic}
\end{algorithm}

\begin{algorithm}[H]
\caption{Descriptor-Conditioned Actor Injection}
\MyLabel{alg:dcrl-me-ai}
\begin{algorithmic}
\Function{\textsc{actor\_injection}}{$\policy_\actorParams$}
    \State $\desc_1, \dots, \desc_{\aibatchsize} \sim \mathcal{U}(\descSpace)$
    \State $\policyParams_1, \dots, \policyParams_{\aibatchsize} \gets G_\actorParams(\desc_1), \dots, G_\actorParams(\desc_{\aibatchsize})$
    \State \Return $\policy_{\policyParams_1}, \dots, \policy_{\policyParams_{\aibatchsize}}$
\EndFunction
\end{algorithmic}
\end{algorithm}

\newpage
\subsection{\dcgme{}}
\begin{algorithm}
\caption{\dcgme{}}
\MyLabel{alg:dcg-me}
\begin{algorithmic}
\Require GA batch size $\gabatchsize$, PG batch size $\pgbatchsize$, Actor Evaluation (AE) batch size $\aebatchsize$, total batch size $b = \gabatchsize + \pgbatchsize$
\State Initialize archive $\archive$ with $b$ random solutions and replay buffer $\replayBuffer$
\State Initialize critic networks $\critic_{\criticParams_1}$, $\critic_{\criticParams_2}$ and actor network $\policy_\actorParams$
\State $i \gets 0$
\While{$i < I$}
    \State $\textsc{train\_actor\_critic}(\policy_\actorParams, \critic_{\criticParams_1}, \critic_{\criticParams_2}, \replayBuffer)$
    \State $\policy_{\policyParams_1}, \dots, \policy_{\policyParams_b} \gets \textsc{selection}(\archive)$
    \State $\policy_{\widehat{\policyParams}_1}, \dots, \policy_{\widehat{\policyParams}_{\gabatchsize}} \gets \textsc{variation\_ga}(\policy_{\policyParams_1}, \dots, \policy_{\policyParams_{\gabatchsize}})$
    \State $\policy_{\widehat{\policyParams}_{\gabatchsize+1}}, \dots, \policy_{\widehat{\policyParams}_b} \gets \textsc{variation\_pg}(\policy_{\policyParams_{\gabatchsize+1}}, \dots, \policy_{\policyParams_b}, \critic_{\criticParams_1}, \replayBuffer)$
    \State $\textsc{addition}(\policy_{\widehat{\policyParams}_1}, \dots, \policy_{\widehat{\policyParams}_b}, \archive, \replayBuffer)$
    \State $i\gets i + b + \aebatchsize$  
\EndWhile

\Function{\textsc{addition}}{$\policy_{\widehat{\policyParams}} \dots, \archive, \replayBuffer$}
    \For{$\desc^\prime \in \descSpace$ sampled from $\aebatchsize$ solutions in $\mathcal{X}$}
        \State $(f, \text{transitions}) \gets F(\policy_\actorParams(\,. \mid \desc^\prime))$
        \State $\textsc{insert}(\mathcal{B}, \text{transitions})$
    \EndFor
    \For{$\policy_{\widehat{\policyParams}} \dots$}
        \State $(f, \text{transitions}) \gets F(\policy_{\widehat{\policyParams}})$, $\desc \gets D(\policy_{\widehat{\policyParams}})$
        \State $\textsc{insert}(\replayBuffer, \text{transitions})$
        \If{$\archive(\desc) = \emptyset$ or $F(\archive(\desc)) < f$}
            \State $\archive(\desc) \gets \policy_{\widehat{\policyParams}}$
        \EndIf
    \EndFor
\EndFunction
\end{algorithmic}
\end{algorithm}

\newpage
\subsection{\pgame{}}
\begin{algorithm}[H]
\caption{\pgame{}}
\label{alg:pga-me}
\begin{algorithmic}
\Require GA batch size $\gabatchsize$, PG batch size $\pgbatchsize$, total batch size $b = \gabatchsize + \pgbatchsize$
\State Initialize archive $\archive$ with $b$ random solutions and replay buffer $\replayBuffer$
\State Initialize critic networks $\critic_{\criticParams_1}$, $\critic_{\criticParams_2}$ and actor network $\policy_\actorParams$
\State $i \gets 0$
\While{$i < I$}
    \State $\textsc{train\_actor\_critic}(\policy_\actorParams, \critic_{\criticParams_1}, \critic_{\criticParams_2}, \replayBuffer)$
    \State $\policy_{\policyParams_1}, \dots, \policy_{\policyParams_{b-1}} \gets \textsc{selection}(\archive)$
    \State $\policy_{\widehat{\policyParams}_1}, \dots, \policy_{\widehat{\policyParams}_{\gabatchsize}} \gets \textsc{variation\_ga}(\policy_{\policyParams_1}, \dots, \policy_{\policyParams_{\gabatchsize}})$
    \State $\policy_{\widehat{\policyParams}_{\gabatchsize+1}}, \dots, \policy_{\widehat{\policyParams}_{b-1}} \gets \textsc{variation\_pg}(\policy_{\policyParams_{\gabatchsize+1}}, \dots, \policy_{\policyParams_{b-1}}, \critic_{\criticParams_1}, \replayBuffer)$
    \State $\policy_{\widehat{\policyParams}_b} \gets \textsc{actor\_injection}(\policy_\actorParams)$
    \State $\textsc{addition}(\archive, \policy_{\widehat{\policyParams}_1}, \dots, \policy_{\widehat{\policyParams}_{b-1}}, \policy_\actorParams, \replayBuffer)$
    \State $i \gets i + b$  
\EndWhile

\Function{\textsc{addition}}{$\archive, \policy_{\widehat{\policyParams}} \dots, \replayBuffer$}
    \For{$\policy_{\widehat{\policyParams}} \dots$}
        \State $(\fitness, \text{transitions}) \gets \fitnessFunction(\policy_{\widehat{\policyParams}})$, $\desc \gets \descFunction(\policy_{\widehat{\policyParams}})$
        \State $\textsc{insert}(\replayBuffer, \text{transitions})$
        \If{$\archive(\desc) = \emptyset$ or $\fitnessFunction(\archive(\desc)) < \fitness$}
            \State $\archive(\desc) \gets \policy_{\widehat{\policyParams}}$
        \EndIf
    \EndFor
\EndFunction
\end{algorithmic}
\end{algorithm}

\begin{algorithm}[H]
\caption{Actor-Critic Training}
\label{alg:pga-me-actor-critic}
\begin{algorithmic}
\Function{\textsc{train\_actor\_critic}}{$\policy_\actorParams, \critic_{\criticParams_1}, \critic_{\criticParams_2}, \replayBuffer$}
    \For{$t = 1  \rightarrow \ncritic{}$}
        \State Sample $N$ transitions $\left(\obs, \action, \reward, \obs^\prime \right)$ from $\replayBuffer$
        \State Sample smoothing noise $\epsilon$
        \State $y \gets \reward + \gamma \min\limits_{i=1,2}  \critic_{\criticParams_{i}^\prime}(s^\prime, \policy_{\actorParams^\prime}(s^\prime) + \epsilon)$
        \State Update both critics by regression to $y$
        \If{$t$ mod $\delay$}
            \State Update actor using the deterministic policy gradient:
            \State 	$\frac{1}{N}\sum \nabla_\actorParams \policy_{\actorParams}(\obs) \nabla_a \critic_{\criticParams_1}(s, \action)|_{\action=\policy_{\actorParams}(s)}$
            \State Soft-update target networks $\critic_{\criticParams{i}^\prime}$ and $\policy_{\actorParams^\prime}$
        \EndIf
    \EndFor
\EndFunction
\end{algorithmic}
\end{algorithm}

\begin{algorithm}[H]
\caption{PG Variation}
\label{alg:pga-me-qpg}
\begin{algorithmic}
\Function{\textsc{variation\_pg}}{$\policy_{\policyParams} \dots,  \critic_{\criticParams_1}, \replayBuffer$}
    \For{$\policy_{\policyParams} \dots$}
        \For{$i = 1  \rightarrow \npg$}
            \State Sample $N$ transitions $\left(\obs, \action, \reward, \obs^\prime \right)$ from $\replayBuffer$
            \State Update actor using the deterministic policy gradient:
            \State 	$\frac{1}{N}\sum \nabla_\policyParams \policy_{\policyParams}(s) \nabla_a \critic_{\criticParams_1}(s, \action)|_{\action=\policy_{\policyParams}(s)}$
        \EndFor
    \EndFor
    \State \Return $\policy_{\widehat{\policyParams}} \dots$
\EndFunction
\end{algorithmic}
\end{algorithm}

\begin{algorithm}
\caption{Actor Injection}
\label{alg:pga-me-ai}
\begin{algorithmic}
\Function{\textsc{actor\_injection}}{$\policy_\actorParams$}
    \State \Return $\policy_\actorParams$
\EndFunction
\end{algorithmic}
\end{algorithm}

\newpage
\subsection{\qdpg{}}
\begin{algorithm}
\caption{\qdpg{}}
\label{alg:qdpg}
\begin{algorithmic}
\Require GA batch size $\gabatchsize$, QPG batch size $\qpgbatchsize$, DPG batch size $\dpgbatchsize$, total batch size $b = \gabatchsize + \qpgbatchsize + \dpgbatchsize$
\State Initialize archive $\archive$ with $b$ random solutions and replay buffer $\replayBuffer$
\State Initialize critic networks $\critic_{\criticParams_Q}$, $\critic_{\criticParams_D}$ and actor network $\policy_\actorParams$
\State $i \gets 0$
\While{$i < I$}
    \State $\textsc{train\_actor\_critic}(\policy_\actorParams, \critic_{\criticParams_Q}, \critic_{\criticParams_D}, \replayBuffer)$
    \State $\policy_{\policyParams_1}, \dots, \policy_{\policyParams_b} \gets \textsc{selection}(\archive)$
    \State $\policy_{\widehat{\policyParams}_1}, \dots, \policy_{\widehat{\policyParams}_{\gabatchsize}} \gets \textsc{variation\_ga}(\policy_{\policyParams_1}, \dots, \policy_{\policyParams_{\gabatchsize}})$
    \State $\policy_{\widehat{\policyParams}_{\gabatchsize+1}}, \dots, \policy_{\widehat{\policyParams}_{\gabatchsize+\qpgbatchsize}} \gets \textsc{variation\_qpg}(\policy_{\policyParams_{\gabatchsize+1}}, \dots, \policy_{\policyParams_{\gabatchsize+\qpgbatchsize}}, \critic_{\criticParams_Q}, \replayBuffer)$
    \State $\policy_{\widehat{\policyParams}_{\gabatchsize+\qpgbatchsize+1}}, \dots, \policy_{\widehat{\policyParams}_b} \gets \textsc{variation\_dpg}(\policy_{\policyParams_{\gabatchsize+\qpgbatchsize+1}}, \dots, \policy_{\policyParams_b}, \critic_{\criticParams_D}, \replayBuffer)$
    \State $\textsc{addition}(\policy_{\widehat{\policyParams}_1}, \dots, \policy_{\widehat{\policyParams}_b}, \archive, \replayBuffer)$
    \State $i\gets i + b$  
\EndWhile

\Function{\textsc{addition}}{$\archive, \replayBuffer, \policy_\actorParams, \policy_{\widehat{\policyParams}} \dots$}
    \For{$\desc^\prime \in \descSpace$ sampled from $b$ solutions in $\archive$}
        \State $(f, \text{transitions}) \gets F(\policy_\actorParams(\,. \mid \desc^\prime))$
        \State $\textsc{insert}(\replayBuffer, \text{transitions})$
    \EndFor
    \For{$\policy_{\widehat{\policyParams}} \dots$}
        \State $(f, \text{transitions}) \gets F(\policy_{\widehat{\policyParams}})$, $\desc \gets D(\policy_{\widehat{\policyParams}})$
        \State $\textsc{insert}(\replayBuffer, \text{transitions})$
        \If{$\archive(\desc) = \emptyset$ or $F(\archive(\desc)) < f$}
            \State $\archive(\desc) \gets \policy_{\widehat{\policyParams}}$
        \EndIf
    \EndFor
\EndFunction
\end{algorithmic}
\end{algorithm}

\newpage
\subsection{\me{}}
\begin{algorithm}[H]
\caption{\me{}}
\label{alg:me}
\begin{algorithmic}
\Require GA batch size $\gabatchsize$
\State Initialize archive $\archive$ with $\gabatchsize$ random solutions
\State $i \gets 0$
\While{$i < I$}
    \State $x_1, \dots, x_{\gabatchsize} \gets \textsc{selection}(\archive)$
    \State $\widehat{x}_1, \dots, \widehat{x}_{\gabatchsize} \gets \textsc{variation}(x_1, \dots, x_{\gabatchsize})$
    \State $\textsc{addition}(\archive, \widehat{x}_1, \dots, \widehat{x}_{\gabatchsize})$
    \State $i \gets i + \gabatchsize$
\EndWhile

\Function{addition}{$\archive, \widehat{x}\dots$} :
    \For{$\widehat{x} \dots$}
        \State $\fitness \gets \fitnessFunction(\widehat{x})$, $\desc \gets \descFunction(\widehat{x})$
        \If{$\archive(\desc) = \emptyset$ or $\fitnessFunction(\archive(\desc)) < \fitness$}
            \State $\archive(\desc) \gets \widehat{x}$
        \EndIf
    \EndFor
\EndFunction
\end{algorithmic}
\end{algorithm}

\newpage
\subsection{\mees{}}
\begin{algorithm}[H]
\caption{\mees{}}
\label{alg:me_es}
\begin{algorithmic}
\Require Number of ES samples $N$, standard deviation of ES samples $\sigma$, explore-exploit alternation $N_{gen}$, number of re-sampling $M$
\State Initialize archive $\archive$ with $N$ random solutions, initialise empty novelty archive $\mathcal{A}$
\State $i \gets 0$
\While{$i < I$}
    \If{$i \% N_{gen} == 0$}:
        \State $x \gets \textsc{selection\_exploit}(\archive)$
        \State $\widehat{x} \gets \textsc{variation\_exploit}(x)$
    \Else:
        \State $x \gets \textsc{selection\_explore}(\archive)$
        \State $\widehat{x} \gets \textsc{variation\_explore}(\mathcal{A}, x)$
    \EndIf
    \State $\textsc{addition}(\archive, \mathcal{A}, \widehat{x})$
    \State $i \gets i + N + M$
\EndWhile

\Function{addition}{$\archive, \mathcal{A}, \widehat{x}$} :
    \For{$i = 1, \dots, M$}
        \State $\fitness_i \gets \fitnessFunction(\widehat{x}))$, $\desc_i \gets \descFunction(\widehat{x}))$
    \EndFor
    \State $\fitness \gets \text{average}(\fitness_i)$, $\desc \gets \text{average}(\desc_i)$
    \State $\mathcal{A} \gets \mathcal{A} + \desc$
    \If{$\archive(\desc) = \emptyset$ or $\fitnessFunction(\archive(\desc)) < \fitness$}
        \State $\archive(\desc) \gets \widehat{x}$
    \EndIf
\EndFunction

\Function{variation\_exploit}{$x$} :
    \State $x_1, \dots, x_{N} \gets \textsc{sample\_gaussian}(x, \sigma)$ 
    \State $\fitness_1, \dots, \fitness_{N} \gets \fitnessFunction(x_1, \dots, x_{N})$
    \State $\widehat{x} \gets \textsc{es\_step}(x, \fitness_1, \dots, \fitness_{N})$
\EndFunction

\Function{variation\_explore}{$\mathcal{A}, x$} :
    \State $x_1, \dots, x_{N} \gets \textsc{sample\_gaussian}(x, \sigma)$ 
    \State $\desc_1, \dots, \desc_{N} \gets \descFunction(x_1, \dots, x_{N})$
    \State $nov_1, \dots, nov_{N} \gets \textsc{novelty}(\mathcal{A}, \desc_1, \dots, \desc_{N})$
    \State $\widehat{x} \gets \textsc{es\_step}(x, nov_1, \dots, nov_{N})$
\EndFunction
\end{algorithmic}
\end{algorithm}

\newpage
\section{Hyperparameters}

\subsection{\dcrlme{}}
\begin{table}[H]
\label{tab:hyperparameters-dcrl-me}
\caption{\dcrlme{} hyperparameters}
\centering
\begin{tabular}{l | c}
\toprule
Parameter & Value\\
\midrule
Number of centroids & $1024$\\
Total batch size $b$ & $256$\\
GA batch size $\gabatchsize$ & $128$\\
PG batch size $\pgbatchsize$ & $64$\\
AI batch size $\aibatchsize$ & $64$\\
Policy networks & [128, 128, $|\mathcal{A}|$]\\
\midrule
GA variation param. 1 $\sigma_1$ & $0.005$\\
GA variation param. 2 $\sigma_2$ & $0.05$\\
\midrule
Actor network & [128, 128, $|\mathcal{A}|$]\\
Critic network & [256, 256, 1]\\
TD3 batch size $N$ & $100$\\
Critic training steps $\ncritic$ & $3000$\\
PG training steps $\npg$ & $150$\\
Policy learning rate & $5 \times 10^{-3}$\\
Actor learning rate & $3 \times 10^{-4}$\\
Critic learning rate & $3 \times 10^{-4}$\\
Replay buffer size & $10^6$\\
Discount factor $\gamma$ & $0.99$\\
Actor delay $\delay$ & $2$\\
Target update rate & $0.005$\\
Smoothing noise var. $\sigma$ & $0.2$\\
Smoothing noise clip & $0.5$\\
\midrule
Length scale $L$ & 0.1\\
\bottomrule
\end{tabular}
\end{table}

\subsection{\dcgme{}}
\begin{table}[H]
\label{tab:hyperparameters-dcg-me}
\caption{\dcgme{} hyperparameters}
\centering
\begin{tabular}{l | c}
\toprule
Parameter & Value\\
\midrule
Number of centroids & $1024$\\
Total batch size $b$ & $256$\\
GA batch size $\gabatchsize$ & $128$\\
PG batch size $\pgbatchsize$ & $128$\\
Policy networks & [128, 128, $|\mathcal{A}|$]\\
\midrule
GA variation param. 1 $\sigma_1$ & $0.005$\\
GA variation param. 2 $\sigma_2$ & $0.05$\\
\midrule
Actor network & [128, 128, $|\mathcal{A}|$]\\
Critic network & [256, 256, 1]\\
TD3 batch size $N$ & $100$\\
Critic training steps $\ncritic$ & $3000$\\
PG training steps $\npg$ & $150$\\
Policy learning rate & $5 \times 10^{-3}$\\
Actor learning rate & $3 \times 10^{-4}$\\
Critic learning rate & $3 \times 10^{-4}$\\
Replay buffer size & $10^6$\\
Discount factor $\gamma$ & $0.99$\\
Actor delay $\delay$ & $2$\\
Target update rate & $0.005$\\
Smoothing noise var. $\sigma$ & $0.2$\\
Smoothing noise clip & $0.5$\\
\midrule
Length scale $L$ & 0.1\\
\bottomrule
\end{tabular}
\end{table}

\subsection{\pgame{}}
\begin{table}[H]
\label{tab:hyperparameters-pga-me}
\caption{\pgame{} hyperparameters}
\centering
\begin{tabular}{l | c}
\toprule
Parameter & Value\\
\midrule
Number of centroids & $1024$\\
Total batch size $b$ & $256$\\
GA batch size $\gabatchsize$ & $128$\\
PG batch size $\pgbatchsize$ & $127$\\
AI batch size $\aibatchsize$ & $1$\\
Policy networks & [128, 128, $|\mathcal{A}|$]\\
\midrule
GA variation param. 1 $\sigma_1$ & $0.005$\\
GA variation param. 2 $\sigma_2$ & $0.05$\\
\midrule
Actor network & [128, 128, $|\mathcal{A}|$]\\
Critic network & [256, 256, 1]\\
TD3 batch size $N$ & $100$\\
Critic training steps $\ncritic$ & $3000$\\
PG training steps $\npg$ & $150$\\
Policy learning rate & $5 \times 10^{-3}$\\
Actor learning rate & $3 \times 10^{-4}$\\
Critic learning rate & $3 \times 10^{-4}$\\
Replay buffer size & $10^6$\\
Discount factor $\gamma$ & $0.99$\\
Actor delay $\delay$ & $2$\\
Target update rate & $0.005$\\
Smoothing noise var. $\sigma$ & $0.2$\\
Smoothing noise clip & $0.5$\\
\bottomrule
\end{tabular}
\end{table}

\subsection{\qdpg{}}
\begin{table}[H]
\label{tab:hyperparameters-qd-pg}
\caption{\qdpg{} hyperparameters}
\centering
\begin{tabular}{l | c}
\toprule
Parameter & Value\\
\midrule
Number of centroids & $1024$\\
Total batch size $b$ & $256$\\
GA batch size $\gabatchsize$ & $86$\\
QPG batch size $\pgbatchsize$ & $85$\\
DPG batch size $\pgbatchsize$ & $85$\\
Policy networks & [128, 128, $|\mathcal{A}|$]\\
\midrule
GA variation param. 1 $\sigma_1$ & $0.005$\\
GA variation param. 2 $\sigma_2$ & $0.05$\\
\midrule
Actor network & [128, 128, $|\mathcal{A}|$]\\
Critic network & [256, 256, 1]\\
TD3 batch size $N$ & $100$\\
Quality critic training steps $\ncritic$ & $3000$\\
Diversity critic training steps $\ncritic$ & $300$\\
PG training steps $\npg$ & $150$\\
Policy learning rate & $5 \times 10^{-3}$\\
Actor learning rate & $3 \times 10^{-4}$\\
Critic learning rate & $3 \times 10^{-4}$\\
Replay buffer size & $10^6$\\
Discount factor $\gamma$ & $0.99$\\
Actor delay $\delay$ & $2$\\
Target update rate & $0.005$\\
Smoothing noise var. $\sigma$ & $0.2$\\
Smoothing noise clip & $0.5$\\
\midrule
Number nearest neighbors & 5\\
Novelty scaling ratio & 1.0\\
\bottomrule
\end{tabular}
\end{table}

\subsection{\me{}}
\begin{table}[H]
\label{tab:hyperparameters-me}
\caption{\me{} hyperparameters}
\centering
\begin{tabular}{l | c}
\toprule
Parameter & Value\\
\midrule
Number of centroids & $1024$\\
Total batch size $b$ & $256$\\
GA batch size $\gabatchsize$ & $256$\\
Policy networks & [128, 128, $|\mathcal{A}|$]\\
\midrule
GA variation param. 1 $\sigma_1$ & $0.005$\\
GA variation param. 2 $\sigma_2$ & $0.05$\\
\bottomrule
\end{tabular}
\end{table}

\subsection{\mees{}}
\begin{table}[H]
\label{tab:hyperparameters-me-es}
\caption{\mees{} hyperparameters}
\centering
\begin{tabular}{l | c}
\toprule
Parameter & Value\\
\midrule
Number of centroids & $1024$\\
Total batch size $b$ & $256$\\
GA batch size $\gabatchsize$ & $128$\\
PG batch size $\pgbatchsize$ & $127$\\
AI batch size $\aibatchsize$ & $1$\\
Policy networks & [128, 128, $|\mathcal{A}|$]\\
\midrule
GA variation param. 1 $\sigma_1$ & $0.005$\\
GA variation param. 2 $\sigma_2$ & $0.05$\\
\midrule
Number of samples & $1000$\\
Sample sigma & $0.02$\\
\bottomrule
\end{tabular}
\end{table}

\end{document}